\pdfoutput=1
\documentclass{article}

\title{Example-Based Image Synthesis via Randomized Patch-Matching}
\author{Yi~Ren, Yaniv~Romano, Michael~Elad\thanks{Y. Ren is with the Computer Science Department, the Technion -- Israel Institute of Technology, Technion City, Haifa 32000, Israel. E-mail address: \texttt{yi.ren@campus.technion.ac.il}. Y. Romano is with the Department of Electrical Engineering, Technion. E-mail address: \texttt{yromano@tx.technion.ac.il}. M. Elad is with the Computer Science Department, Technion. E-mail address: \texttt{elad@cs.technion.ac.il}.}}

\date{\today}

\usepackage{cite}
\usepackage{amsmath}
\usepackage{amsfonts}
\usepackage{mathtools}

\usepackage{graphicx}
\usepackage{url}
\usepackage[bookmarks=true,hidelinks]{hyperref}

\usepackage{tablefootnote}
\usepackage{color}

\usepackage[]{algorithm2e}

\usepackage{framed}

\DeclarePairedDelimiter{\abs}{\lvert}{\rvert}
\DeclarePairedDelimiter{\norm}{\lVert}{\rVert}

\DeclareMathOperator*{\argmin}{arg\,min}
\DeclareMathOperator*{\NN}{NN}
\DeclareMathOperator{\Tr}{Tr}
\DeclareMathOperator{\rank}{rank}

\begin{document}

\maketitle

\begin{abstract}
Image and texture synthesis is a challenging task that has long been drawing attention in the fields of image processing, graphics, and machine learning. This problem consists of modelling the desired type of images, either through training examples or via a parametric modeling, and then generating images that belong to the same statistical origin.

This work addresses the image synthesis task, focusing on two specific families of images -- handwritten digits and face images. This paper offers two main contributions. First, we suggest a simple and intuitive algorithm capable of generating such images in a unified way. The proposed approach taken is pyramidal, consisting of upscaling and refining the estimated image several times. For each upscaling stage, the algorithm randomly draws small patches from a patch database, and merges these to form a coherent and novel image with high visual quality. The second contribution is a general framework for the evaluation of the generation performance, which combines three aspects: the likelihood, the originality and the spread of the synthesized images. We assess the proposed synthesis scheme and show that the results are similar in nature, and yet different from the ones found in the training set, suggesting that true synthesis effect has been obtained.

\end{abstract}

\section{Introduction} \label{sec:introduction}

The task of image synthesis is central in the fields of image processing, graphics, and machine learning. The motivation to study this topic has several origins. First, the availability of a technique for generating images that obey a given patch-based image model establishes an ultimate way for testing local models and their suitability to treat images. Second, a successful synthesis algorithm may improve the performance of image restoration tasks, e.g. denoising \cite{elad2006image,dabov2007image,mairal2008sparse}, demosaicking \cite{mairal2008sparse}, inpainting \cite{criminisi2004region}, super-resolution \cite{yang2010image}, and other ill-posed inverse problems \cite{zoran2011learning}, as well as related image processing tasks, such as image analogies\cite{hertzmann2001image,cheng2008consistent,benard2013stylizing,barnes2015patchtable} or style transfer\cite{kyprianidis2013state}. Moreover, in case of severe corruption, the conventional image restoration algorithms do not achieve clear and sharp images, while such synthesis methods could be used to provide a plausible solution, out of infinitely many possible ones. Put differently, rather than considering a solution as a direct recovery task, armed with a good synthesis method one may migrate the treatment to have a randomized hallucination flavor. Third, the generated images themselves are interesting, since they are created out of ``thin air", and may be considered as an appealing art. Indeed, this is the effect of the recently introduced synthesized images \cite{kyprianidis2013state,gatys2016image,johnson2016perceptual,bruna2015super,dong2016image,radford2015unsupervised,van2016pixel,gregor2015draw,kingma2016improving,salimans2016improved,google2015inceptionism}. Finally, the synthesis methods can be converted to handle other data sources such as music, thereby enriching the scope of this field of research far beyond its original objectives.

When addressing the general image synthesis problem, one may narrow down the scope to a more specific task to ease the otherwise quite complicated general objective. Indeed, over the years many works focus on texture synthesis \cite{portilla2000parametric,peyre2010texture,peyre2009sparse,tartavel2014variational,paget1995texture,efros2001image,kwatra2003graphcut,kwatra2005texture,gatys2015texture,ashikhmin2001synthesizing,lefebvre2005parallel,lefebvre2006appearance,efros1999texture}, handwritten digits \cite{salakhutdinov2008quantitative,uria2014deep,raiko2014iterative,salakhutdinov2009deep,murray2009evaluating,rezende2014stochastic,salimans2015markov,gregor2014deep,gregor2015draw,radford2015unsupervised,van2016pixel,kingma2016improving,hinton2006fast,goodfellow2014generative,kim2016deep,germain2015made}, and human faces \cite{ranzato2013modeling,rezende2014stochastic,larsen2015autoencoding,radford2015unsupervised,goodfellow2014generative,kim2016deep,liu2007face}. These cases are all appealing as test cases for synthesis algorithms, and as a stepping stone towards the ultimate problem of general content image generation. This is also the path we shall take in this work.

Traditionally, in the case of texture synthesis, the generation is regularly done using example-based models \cite{paget1995texture,efros2001image,kwatra2003graphcut,kwatra2005texture,ashikhmin2001synthesizing,lefebvre2005parallel,lefebvre2006appearance,efros1999texture}. Newly emerging methods in the field of image synthesis commonly model the probability distribution of the images by neural networks, then randomly sample from it to generate new content \cite{denton2015deep,mahendran2015understanding,simonyan2013deep,dosovitskiy2015inverting,nguyen2015deep,hinton2006fast,ranzato2013modeling,gregor2015draw,van2016pixel,larsen2015autoencoding,gatys2016image,johnson2016perceptual,goodfellow2014generative,kim2016deep,gatys2015texture,radford2015unsupervised,kingma2016improving,salakhutdinov2008quantitative,uria2014deep,raiko2014iterative,salakhutdinov2009deep,murray2009evaluating,rezende2014stochastic,salimans2015markov,gregor2014deep,germain2015made}. These methods have been shown to lead to interesting results for generating digits, faces, textures, and even natural scenes. However, a fundamental drawback of these methods is their tendency to be over-complicated and difficult to interpret. 

Interestingly, in the realm of image synthesis, little existing work has relied on patch-based image models, despite the demonstrated effectiveness of such techniques in image restoration \cite{elad2006image,dabov2007image,mairal2008sparse,criminisi2004region,yang2010image,zoran2011learning} and texture-synthesis \cite{kwatra2005texture} tasks. More specifically, modeling of patches, either directly using examples\cite{freeman2002example,freeman2000learning,criminisi2004region,efros2001image,ashikhmin2001synthesizing,lefebvre2005parallel,lefebvre2006appearance,efros1999texture,kwatra2005texture,kwatra2003graphcut} or via a parametric form \cite{yu2012solving,zoran2011learning,mairal2009non}, has been shown to be highly expressive and rich. Perhaps the reason of avoiding such localized methods in image synthesis is the inevitable need to quilt or otherwise aggregate these patches to form the final created image in a way that is globally faithful.

A restoration problem that is closely related to synthesis is the single image super-resolution (or upscaling), where the goal is to increase the resolution of a given degraded image such that the restored image would be as close as possible to the original one. This highly ill-posed problem has been treated quite effectively using patch-based methods \cite{freeman2000learning,freeman2002example,yang2010image,glasner2009super,timofte2014a+,papyan2016multi,romano2014single}, which essentially inject and hallucinate new content in a realistic fashion to form the super-resolved outcome. As an example for such a work, a popular super-resolution algorithm is the one reported in \cite{freeman2000learning,freeman2002example}, relying on an example-based patch model. Their approach constructs a database of low-resolution (LR) and high-resolution (HR) patch pairs, and super-resolves a LR image by matching the patches of the input image to the LR part of the database, this way finding high-resolution candidate patches to replace the input. This is then followed by a Belief-Propagation based stitching process of the HR patches to form the desired high-resolution result.

The EPLL \cite{zoran2011learning} is another successful patch-based image restoration algorithm. Rather than using examples directly, this method builds upon a parametric local Gaussian Mixture Model (GMM). Another key difference with respect to \cite{freeman2002example,freeman2000learning} is the patch-fusion process. EPLL promotes the patches of the final restored image to comply with the local model, by alternating between (1) restoring each patch taken from the previous estimate according to the local prior, and (2) reconstructing the image by applying a patch-averaging step. This iterative process results in a significant reduction of artifacts, leading to state-of-the-art restoration. Incidentally, a similar concept to EPLL has been presented in \cite{kwatra2005texture}, in the context of texture synthesis. More specifically, while \cite{kwatra2005texture}
 relies directly on example patches from a given texture image (rather than a parametric model), its main objetive and way to achieve it are the same as in the EPLL: The goal is getting a synthesized image in which every patch has a close match to the example set, and the way to get this is by a similar iterative process.
 
In this paper we propose to leverage the example-based model and the EPLL framework, putting forward a multi-scale image generation algorithm, which is intuitive and competitive with the state-of-the-art, while being fully interpretable. Given a very small seed image (e.g. $4 \times 4$ pixels) as an input, we first upscale it by factor of 2 in each axis using a simple interpolation method (e.g. bilinear). Then, new content is \emph{hallucinated} via an example-based spatially varying local priors, which are plugged to the EPLL scheme. Clearly, this process can be repeated several times, until the image reaches the desired size. Following previous work \cite{salakhutdinov2008quantitative,uria2014deep,raiko2014iterative,salakhutdinov2009deep,murray2009evaluating,rezende2014stochastic,salimans2015markov,gregor2014deep,gregor2015draw,radford2015unsupervised,van2016pixel,kingma2016improving,hinton2006fast,goodfellow2014generative,kim2016deep,ranzato2013modeling,larsen2015autoencoding,liu2007face,germain2015made}, we test the ability of the proposed algorithm to synthesize images of both handwritten digits (based on MNIST \cite{lecun1998gradient}) and human faces, showing that we achieve results comparable to the state-of-the-art. Our proposed algorithm bares some similarities to the texture-synthesis work reported in \cite{kwatra2005texture}. We shall come back to this matter and map clearly the differences between \cite{kwatra2005texture} and our algorithm.

Our second contribution is a framework for assessing the performance of an arbitrary synthesis machine. Previous work has evaluated the performance of the synthesis algorithm by computing the Log-Likelihood (LL) of the test images in the probability distribution associated with their parametric synthesizer \cite{salakhutdinov2008quantitative,uria2014deep,raiko2014iterative,salakhutdinov2009deep,murray2009evaluating,rezende2014stochastic,salimans2015markov,gregor2014deep,gregor2015draw,van2016pixel,kingma2016improving,goodfellow2014generative,ranzato2013modeling,germain2015made}. This measure indeed indicates the generalization power of the trained model. However, since the test set contains only real images, it does not necessarily assign low probabilities to undesired images (such as the blank image and ones containing artifacts). Thus, the LL value obtained on the test images does not indicate whether such failed images may be generated. Therefore to evaluate the visual quality of the synthesis outcome, it is indispensable to assess the LL directly on the generated images, in a manner that is non-parametric or independent from the synthesis model.

However, moving from LL evaluation of the test images to the LL of synthesized ones is not sufficient, as this measure does not reveal the whole picture. Suppose we have a trivial synthesis machine that memorized  training images and provide them as its output. In this case, the generated images are of high LL measure, and they are \emph{spread} evenly over the training images, but \emph{without introducing any new content}. Furthermore, consider another machine producing always the same image which is both novel and of high quality. Here the synthesized images have better \emph{originality}, but are concentrated on a single point, ignoring the distribution of the training images. In these two cases the LL will indicate an excellent performance while the synthesis is both deterministic and degenerated. Motivated by this observation, we propose a complete assessment framework combining three aspects of the performance: the likelihood, the originality and the spread of the synthesized images. Both numerical measures and visualization tools are presented to evaluate these three aspects. Our experiments show that the proposed synthesis algorithm results in high quality images with good likelihood and originality, along with a reasonable spread.

This paper is organized as follows: In Section \ref{sec:epll} we review the EPLL algorithm as we will rely on it in the proposed synthesis scheme, which is described in Section \ref{sec:algorithm} with all its ingredients. Then, in Section \ref{sec:synthesis experiments} we provide various synthesis results of digits and faces that demonstrate the effectiveness of the proposed scheme. Section \ref{sec:assessment} presents our way for assessing the goodness of the results, and provides the assessment results of our synthsis outcome, compared to the state-of-the-art. In Section \ref{sec:discussion} we summarize the paper and outline future research directions. 

\section{Background: EPLL via ADMM}
\label{sec:epll}

% The synthesis task can be tackled by applying recursively a randomized super-resolution solver, such as the one in \cite{freeman2002example}, to a small seed image. Nevertheless, in \cite{freeman2002example} the patches are treated in a given order, meaning that patches appearing earlier have greater privilege. To deal with the patches equally, we decide to design our synthesis scheme based on EPLL \cite{zoran2011learning}, in which all individual patches are restored at the same time, then taken into account altogether for the image reconstruction.

In this section we present the EPLL with a slightly changed form --- rather than relying on the quadratic half-splitting strategy used in \cite{zoran2011learning}, we base our derivations on the more accurate Alternating Directions Methods of Multipliers (ADMM) method \cite{boyd2011distributed} which has been shown to be very effective in numerous applications.

% We start with few words about the notation used in this paper: Uppercase variables are used for whole images (e.g. $X, Y$) and lowercase ones for patches or set of patches (e.g. $u_i^k, z$). Underlined variables indicate that they are converted from an image to a vector form (e.g. $\underline{X}, \underline{z}$). Bold Capital letters indicate operators such as blur or down-sampling.

The core idea behind the EPLL is to regularize the given inverse problem by specifying a prior on patches only. The regularization term averages over the individual patch priors in a form of an expectation, explaining the name given to this method. The restoration problem suggested by EPLL can be formulated as follows:

\begin{equation} \label{eq:epll}
\min_{\underline{X}} \left\{\frac{\lambda}{2}\norm{\mathbf{H}\underline{X} - \underline{Y}}_2^2 - \sum_{i \in I}\log P_i(\mathbf{R}_i \underline{X})\right\}
\end{equation}
\noindent where

\begin{itemize}
    \item $\mathbf{H}$ is the matrix representing the degradation operator.
    \item $Y$ is the input image, which is assumed to be corrupted by $\mathbf{H}$.
    \item $X$ is the output restored image of size $w \times h$.
    \item $I$ is the set of locations of the fully overlapped patches of size $n \times n$ in image $X$.
    \item $\mathbf{R}_i \in \mathbb{R}^{n^2 \times (w \cdot h)}$ is a patch extraction operator for each patch location $i$, e.g. $\mathbf{R}_i \underline{X}$ is the $n^2$-dimensional patch of $X$ at location $i$ (being reordered as a vector).
    \item $P_i(\underline{z}): \mathbb{R}^{n^2} \rightarrow [0,1]$ represents the prior probability function of a patch $z$ located at position $i$. As we use local patch models, this prior may be space- (and scale-) dependent.
\end{itemize}

\noindent The optimization problem in Equation ($\ref{eq:epll}$) is equivalent to

\begin{equation} \label{eq:epll split}
\min_{\underline{X}, \underline{z}} \left\{ \frac{\lambda}{2}\norm{\mathbf{H}\underline{X} - \underline{Y}}_2^2 - \sum_{i \in I}\log P_i(\underline{z}_i)\right\} \text{ s.t. } \mathbf{R}_i \underline{X} - \underline{z}_i = 0 \;(\forall i \in I),
\end{equation}
\noindent where $\underline{z} = \{ \underline{z}_i\}_{i \in I}$ are the auxiliary variables representing the restored patches. Using ADMM, the above problem can be rewritten as

\begin{equation} \label{eq:epll admm}
\min_{\underline{X}, \underline{z}, \{\underline{u}_i\}_{i \in I}} \left\{ \frac{\lambda}{2}\norm{\mathbf{H}\underline{X} - \underline{Y}}_2^2 - \sum_{i \in I}\log P_i(\underline{z}_i) + \frac{\rho}{2}\sum_{i \in I} \norm{\mathbf{R}_i \underline{X} + \underline{u}_i - \underline{z}_i}_2^2 \right\}
\end{equation}
\noindent in which $\underline{u}_i$ plays the role of a Lagrange multiplier for the $i^{th}$ constraint (i.e. $\mathbf{R}_i \underline{X} - \underline{z}_i = 0$), and $\rho$ is the weight of the corresponding penalties. This problem can be solved by the following iterative and alternating steps:

\begin{itemize}
    \item $z$-step:
    \begin{equation} \label{eq:epll z step}
        \forall i \in I, \underline{z}_i^{k+1} = \argmin_{\underline{z}} \left\{-\log P_i (\underline{z}) + \frac{\rho}{2}\norm{\mathbf{R}_i \underline{X}^k + \underline{u}_i^k - \underline{z}}_2^2\right\}.
    \end{equation}
    \noindent This local step is formulated as a MAP estimation of $\underline{z}_i$ given the measurement vector $\mathbf{R}_i \underline{X}^k + \underline{u}_i^k$.
    \item $X$-step:
    \begin{equation} \label{eq:epll x step}
    \begin{split}
    \underline{X}^{k+1} &= \argmin_{\underline{X}} \left\{\frac{\lambda}{2}\norm{\mathbf{H} \underline{X} - \underline{Y}}_2^2 + \frac{\rho}{2}\sum_{i \in I} \norm{\mathbf{R}_i \underline{X} - \underline{z}_i^{k+1} + \underline{u}_i^{k}}_2^2\right\} \\
    &= \Big(\lambda \mathbf{H}^T \mathbf{H} + \rho \sum_{i \in I} \mathbf{R}_i^T \mathbf{R}_i\Big)^{-1} \Big(\lambda \mathbf{H}^T \underline{Y} + \rho \sum_{i \in I} \mathbf{R}_i^T(\underline{z}_i^{k+1} - \underline{u}_i^k)\Big).
    \end{split}
    \end{equation}
    \noindent This step merges the estimated patches together to form the global image, while taking into consideration the corrupted image $Y$ and the degradation operator $\mathbf{H}$. This step is essentially built upon a patch averaging procedure, coupled with a Wienner restoration \cite{zoran2011learning,aharon2006k}.
    \item $u$-step:
    \begin{equation} \label{eq:epll u step}
        \forall i\in I, \underline{u}_i^{k+1} = \underline{u}_i^{k} + (\mathbf{R}_i \underline{X}^{k+1} - \underline{z}_i^{k+1}),
    \end{equation}
    \noindent which updates the Lagrange multipliers vectors according to the Augmented Lagrange method\cite{boyd2011distributed}.
\end{itemize}
\noindent We should note that \cite{zoran2011learning} chose to apply GMM for the patch prior $P_i$ at all locations $i \in I$, while in this work we will use an example-based prior --- see Section \ref{subsec:local nn patch prior} for more details.

\section{Patch-Based Image Synthesis Algorithm}
\label{sec:algorithm}

The core challenge of applying directly an existing image restoration scheme for synthesis is the lack of randomness in the output, which we overcome by a multi-scale synthesis scheme described in Section \ref{subsec:restoration to synthsis}, in which a super-resolution via randomized EPLL is applied iteratively. Moreover, it is crucial to choose a traceable patch prior in the EPLL in order to sample sharp and likely patches. We propose for this need a non-parametric example-based prior in Section \ref{subsec:local nn patch prior}. Other issues also arise, for example:
\begin{itemize}
    \item The patch size: An inherent property of our patch-based method is the locality, which is more intuitive and computationally more tractable compared to the use of global models. However, a dilemma is originated from the choice of the patch size: small patches do not capture wide range information in the images, while large patches harm the richness of the generation. This is solved by the multi-scale structure of our method and the patch context extension we use.
    \item The patch overlap: The overlap between patches is associated with a trade-off between the visual quality and the richness of the resulting images, which is bypassed by using the cycle-spinning extension\cite{coifman1995wavelets,ram2013image}.
\end{itemize}
These extensions are presented in Section \ref{subsec:extensions}.

% Our pyramidal approach of synthesis can be split into two smaller tasks: (1) given a low resolution image $Y$, find a high resolution image $X$ such that the down-scaling of $X$ is close to $Y$, and each patch from $X$ is a likely patch (algorithm \ref{alg:synthesis single}); (2) apply algorithm \ref{alg:synthesis single} iteratively to a very small image to generate a large image in a multi-scale fashion (algorithm \ref{alg:synthesis multi-scale}).

% Before going into the synthesis algorithm, we need to review the original EPLL algorithm, which is our starting point, in its slightly changed form (Section \ref{subsec:EPLL review}); then we show the modifications to EPLL that lead us to algorithm \ref{alg:synthesis single} and their rationale (Section \ref{subsec:restoration to synthsis}); next we describe the algorithm 2 in Section \ref{subsec:synthesis multi-scale}; Finally, we introduce the prior we will be using in our synthesis experiments in Section \ref{subsec:local nn patch prior}.

\subsection{From Restoration to Synthesis}
\label{subsec:restoration to synthsis}
In this subsection we turn to develop the proposed synthesis algorithm. The approach taken is pyramidal, where we suggest repeating the following process: Given a low resolution image $Y$, we first upscale it by factor 2 in each axis using a simple interpolation method, leading to the image $X$. Then, the randomized EPLL is utilized to refine the estimation $X$ by promoting its patches to comply with a local \emph{spatially varying} patch-prior, while being close to the LR image $Y$. Meanwhile, randomness force is injected into the EPLL (as described below) in order to avoid deterministic results, in contrast to the conventional super-resolution methods. The above upscaling process, termed {\em Layer-Synthesis}, can be repeated several times, starting from an extremely small image containing almost no information (such an image is referred to hereafter as the \emph{seed}), and leading to an image of any desired size, as long as we have adequate local priors to drive the process.

We now consider the question of how to convert a super-resolver into a synthesis algorithm with randomness. Given a LR image $Y$ and a down-sampling operator\footnote{For simplicity, we assume that $\mathbf{H}$ halves the size of the input image in each axis throughout the paper.}$\mathbf{H}$, the goal of both patch-based super-resolution and synthesis is to obtain a HR image $X$ such that $\mathbf{H}\underline{X}$ is close to  the input $\underline{Y}$ and each patch from $\underline{X}$ is likely under a local patch model. Oftentimes, this process is formulated as an optimization problem, as described in Equation (\ref{eq:epll}), which can be solved efficiently using the EPLL. However, in contrast to the deterministic outcome of a super-resolver, a synthesis method is also expected to generate many different HR images, all being plausible high-resolution versions of $Y$. Thus, we should allow randomness in our synthesis algorithm in order to diversify the output possibilities. To this end, we change the patch estimation stage in the EPLL (refer to Equation (\ref{eq:epll z step}) in Section \ref{sec:epll} -- the $z$-step), so that instead of minimizing the following function:
$$
-\log P_i (\underline{z}) + \frac{\rho}{2}\norm{\mathbf{R}_i \underline{X}^k  + \underline{u}_i^k - \underline{z}}_2^2,
$$
we randomly draw a patch from the posterior distribution of $\underline{z}$, whose density is given by
\begin{equation} \label{eq:patch posterior}
T(\underline{z} | i, \underline{X}^k, \underline{u}_i^k) = \frac{1}{G_i} \cdot \exp \left\{\log P_i (\underline{z}) - \frac{\rho}{2}\norm{\mathbf{R}_i \underline{X}^k + \underline{u}_i^k - \underline{z}}_2^2\right\},
\end{equation}
\noindent where $G_i$ is the partition function, $P_i$ is our example-based patch prior which is described in detail in Section \ref{subsec:local nn patch prior}, and $\underline{u}_i^k$ is a scaled dual variable being a by-product of the ADMM. The first term in the exponent encourages the generated patch $\underline{z}$ to align with the local prior by preferring higher $P_i(\underline{z})$ values. The second term enforces $\underline{z}$ to fit to the current estimate of the patch being refined $\mathbf{R}_i \underline{X}^k + \underline{u}_i^k$.

Plugging this randomness force to the EPLL scheme leads to the proposed \textit{Layer-Synthesis} algorithm, which is summarized in Algorithm \ref{alg:synthesis single}. Armed with this single scale generation step, a seed image can be gradually upscaled by invoking the {\em Layer-Synthesis} several times, formulating our multi-scale approach as described in Algorithm \ref{alg:synthesis multi-scale}.

Notice that the proposed method does not create totally arbitrary images as all emerge from a given seed image $Y$. Using a fixed seed for different synthesis runs, we aim to show the randomness power of our method. Furthermore, as the seed images are very small ($4 \times 4$ for digits, $8 \times 8$ for faces), the images created are expected to be quite diverse. As a side note we mention that we could simply model the seed images by Gaussian or GMM and draw them randomly. However, in this work we choose to use seeds taken from known test images, in order to be able to compare the outcomes to the original test image they correspond to.

\begin{framed}
\begin{algorithm}[H]
\SetKwInput{Input}{Given Data}
\SetKwInput{Output}{Output}
\SetKwInput{Initialization}{Initialization}
 \Input{\begin{itemize}
     \setlength\itemsep{0.1em}
     \item $Y$: The LR image
     \item $\mathbf{H}$: The 2D 2:1 down-sampling operator
     \item $n$: The height and width of a patch
     \item $I$: Set of fully overlapping patch locations
    %  \item $\{\text{offset}_k\}_{k=0}^{K-1}$: Offsets of patch locations for cycle spinning
     \item $\{P_i\}_{i}$: The patch priors for all patch locations $i$
     \item $K$: The number of iterations of the ADMM
     \item $\{\lambda_k, \rho_k\}_{k = 0}^{K-1}$: EPLL Parameters
 \end{itemize}}
    \Output{
        \begin{itemize}
            \item Twice higher-resolution synthesized image $X$
        \end{itemize}
    }
    \Initialization{
        \begin{itemize}
            \setlength\itemsep{0.1em}
            \item Set $\underline{u}_i^0 = 0$ for $\forall i \in I$\;
            \item Obtain $\underline{X}^0$ by an interpolation (e.g. bilinear) of $Y$\;
        \end{itemize}
    }
    \For{$k = 0:1:K-1$}{
        \begin{itemize}
            \setlength\itemsep{0.1em}
            % \item Shift the patch locations: $I_k = I + \text{offset}_k$\;
            \item For all patch locations $i \in I$, draw a patch $\underline{z}_i^{k+1}$ at random\\ from the conditional distribution given in Equation (\ref{eq:patch posterior})\;
            \item Reconstruct the image $\underline{X}^{k+1}$ from \\ $\{\underline{z}_i^{k+1}\}_{i \in I}$, according to Equation (\ref{eq:epll x step})\;
            \item For all $i \in I$, update $\underline{u}_i^{k+1}$ using Equation (\ref{eq:epll u step})\;
        \end{itemize}
    }
    $X = X^{K-1}$\;
    \vspace{0.2in}
 \label{alg:synthesis single}
 \caption{$\underline{X} = \textit{Layer-Synthesis}(\underline{Y},\{P_i\}_{i \in I})$}
\end{algorithm}
\end{framed}

%  $\underline{X} = \textit{Layer-Synthesis}(\underline{Y},\mathbf{H},n,I,\{\text{offset}_k\}_{k=0}^{K-1},\{P_i\}_{i \in I},K,\{\lambda_k, \rho_k\}_{k = 0}^{K-1})$

 \begin{framed}
\begin{algorithm}[H]
\SetKwInput{Input}{Given Data}
\SetKwInput{Output}{Output}
\SetKwInput{Initialization}{Initialization}
 \Input{\begin{itemize}
     \setlength\itemsep{0.1em}
     \item $Y$: The seed image
     \item $L$: The number of up-scalings to perform
     \item $\{\mathbf{H}_l\}_{l=0}^{L-1}$: The down-sampling operators
     \item  $n$: The height and width of a patch
     \item $\{I_l\}_{l=0}^{L-1}$: Fully overlapping patch locations at each layer
    %  \item $\{\{\text{offset}_{k,l}\}_{k=0}^{K_l-1}\}_{l=0}^{L-1}$: Offsets of patch locations for cycle spinning 
     \item $\{P_{i,l}\}_{i, 0 \leq l \leq L-1}$: The patch priors per scale and location
     \item $\{K_l\}_{l = 0}^{L-1}$: The number of iterations
     \item $\{\lambda_{k,l}, \rho_{k,l}\}_{0 \leq k \leq K_l - 1, 0 \leq l \leq L-1}$: EPLL Parameters 
 \end{itemize}}
 \Output{\begin{itemize}
     \item HR synthesized image $\underline{X}_0$, up-scaled $L$ times from $\underline{Y}$
 \end{itemize}}
    \Initialization{\begin{itemize}
        \item $\underline{X}_L = \underline{Y}$\;
    \end{itemize}}
    \For{$l = L-1:-1:0$}{
        $\underline{X}_l = \textit{Layer-Synthesis}(\underline{X}_{l+1}, \{P_{i,l}\}_{i})$\;
        % $\underline{X}^l = \textit{Layer-Synthesis}(\underline{X}^{l+1}, \mathbf{H}_{l}, n, I_l, \{\text{offset}_{k,l}\}_{k=0}^{K_l-1}, \{P_{i,l}\}_{i \in I_l}, K_l, \{\lambda_{k,l}, \rho_{k,l}\}_{k = 0}^{K_l-1})$\;
    }
    \vspace{0.2in}
 \label{alg:synthesis multi-scale}
 \caption{Multi-scale synthesis algorithm}
\end{algorithm}
\end{framed}

\subsection{Local Nearest-Neighbor Patch Prior}
\label{subsec:local nn patch prior}

% maybe still useful
% Over the years, numerous patch priors have been developed, such as sparsity-based ones \cite{elad2006image,dabov2007image}, GMM \cite{zoran2011learning,yu2012solving}, low-rank-based models \cite{kwok2015colorization,schaeffer2013low,hu2015patch,lu2014depth,dong2014nonlocal}, to name a few. While each of these models could be plugged into our scheme, we favor the example-based prior. This choice is motivated by the success of \cite{kwatra2005texture} in the task of texture synthesis and the simplicity of this prior, as it is a non-parametric one that relies on the availability of large set of example images.

We now turn to construct the proposed spatially varying example-based synthesis priors, which are plugged to Equation (\ref{eq:patch posterior}). These are formulated as a collection of LR-HR example patch pairs, assembled in dictionaries $D_{LR}^{i, l}$ and $D_{HR}^{i, l}$ for each patch location $i$ and each layer $l$. The construction of these is elucidated in Section \ref{build pair dictionary}. Next, in Section \ref{sampling from dictionaries}  we describe the sampling process from the prior using $D_{LR}^{i, l}$ and $D_{HR}^{i, l}$.

\subsubsection{Building Example Dictionaries} \label{build pair dictionary}
\begin{figure}
    \centering
    \includegraphics[width=0.9\textwidth]{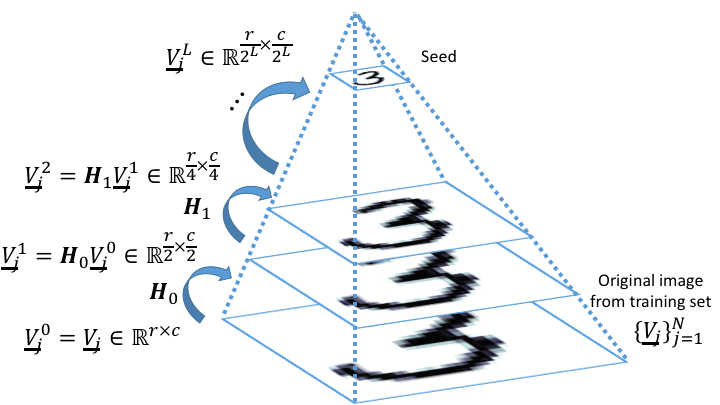}
    \caption{The building process of the Gaussian pyramid for the image $V_j$ from the set of training images.}
    \label{fig:gaussian pyramid}
\end{figure}

In order to construct the LR-HR patch pair dictionaries $D_{LR}^{i, l}$ and $D_{HR}^{i, l}$, we decompose the training images $\{V_j\}_{j=1}^N$ ($V_j \in \mathbb{R}^{r \times c}$) into Gaussian pyramids \cite{forsyth2012computer} (see Figure \ref{fig:gaussian pyramid}), thereby creating a sequence of layers $V_j^0, V_j^1, \ldots V_j^L$ where 
$$\underline{V}_j^0 = \underline{V}_j\; \text{ (original image) }$$ 
\noindent and 
$$\underline{V}_j^l = \mathbf{H}_{l-1} \underline{V}^{l-1}_j, l=1, \ldots L.$$
\noindent The matrices $\mathbf{H}_0,\ldots,\mathbf{H}_{L-1}$ are down-sampling operators (which apply blur and 2:1 decimation in each axis), as such, each layer $V_j^l \in \mathbb{R}^{(r/2^l)\times (r/2^l)}$ is twice smaller than the previous one $V_j^{l-1} \in \mathbb{R}^{(r/2^{l-1})\times (r/2^{l-1})}$. Following this rationale, the image $V_j^L \in \mathbb{R}^{(r/2^L) \times (c/2^L)}$ is the smallest one, corresponding to the seed size.

Next, $D_{LR}^{i, l}$ for layer $l$ and location $i$ is the set of LR patches originating from the lower resolution images $\{V^{l+1}_j\}_{j=1}^N$, given by
$$
D_{LR}^{i, l} = \{\mathbf{Q}_{\frac{i}{2},l+1}\underline{V}_j^{l+1}\}_{j=1}^N,
$$
\noindent where, following Figure \ref{fig:extract patch}, $\mathbf{Q}_{\frac{i}{2},l+1}$ extracts the patch of size $(n/2) \times (n/2)$ from location $\frac{i}{2}$ of the images $\{V^{l+1}_j\}_{j=1}^N$. Similarly, the set $D_{HR}^{i,l}$ contains the HR patches corresponding to the elements in $D_{LR}^{i, l}$, which are extracted from the larger (higher resolution) images $\{V^{l}_j\}_{j=1}^N$. This can be formulated by
$$
D_{HR}^{i, l} = \{\mathbf{R}_{i,l}\underline{V}_j^l\}_{j=1}^N,
$$
\noindent where $\mathbf{R}_{i,l}$ extracts the $i$-th patch of size $n \times n$ from $\{V^l_j\}_{j=1}^{N}$, as depicted in Figure \ref{fig:extract patch}.

\begin{figure}
    \centering
    \includegraphics[width=0.9\textwidth]{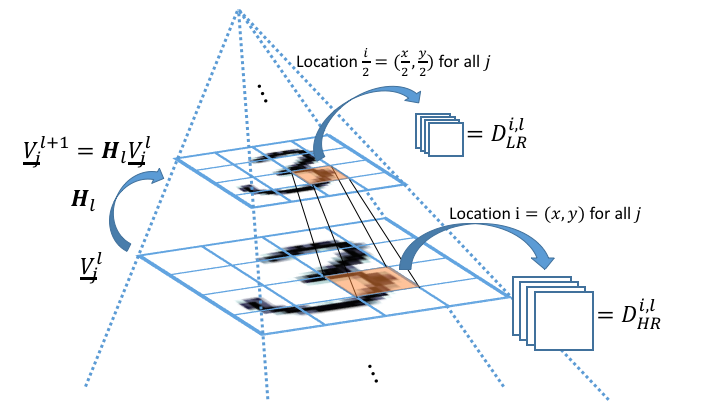}
    \caption{The building process of the prior $P_{i, l}$ for scale $l$ and for each patch location $i \in I_l$. In the figure $i=(x,y)$ means $(x,y)$ is the central pixel of the HR patch at location $i$, and $\frac{i}{2} = (\frac{x}{2}, \frac{y}{2})$ means  $(\frac{x}{2}, \frac{y}{2})$ is the central pixel of the corresponding LR patch at location $\frac{i}{2}$.}
    \label{fig:extract patch}
\end{figure}

\subsubsection{Sampling Process in Synthesis} \label{sampling from dictionaries}
\begin{figure}
    \centering
    \includegraphics[width=1.0\textwidth]{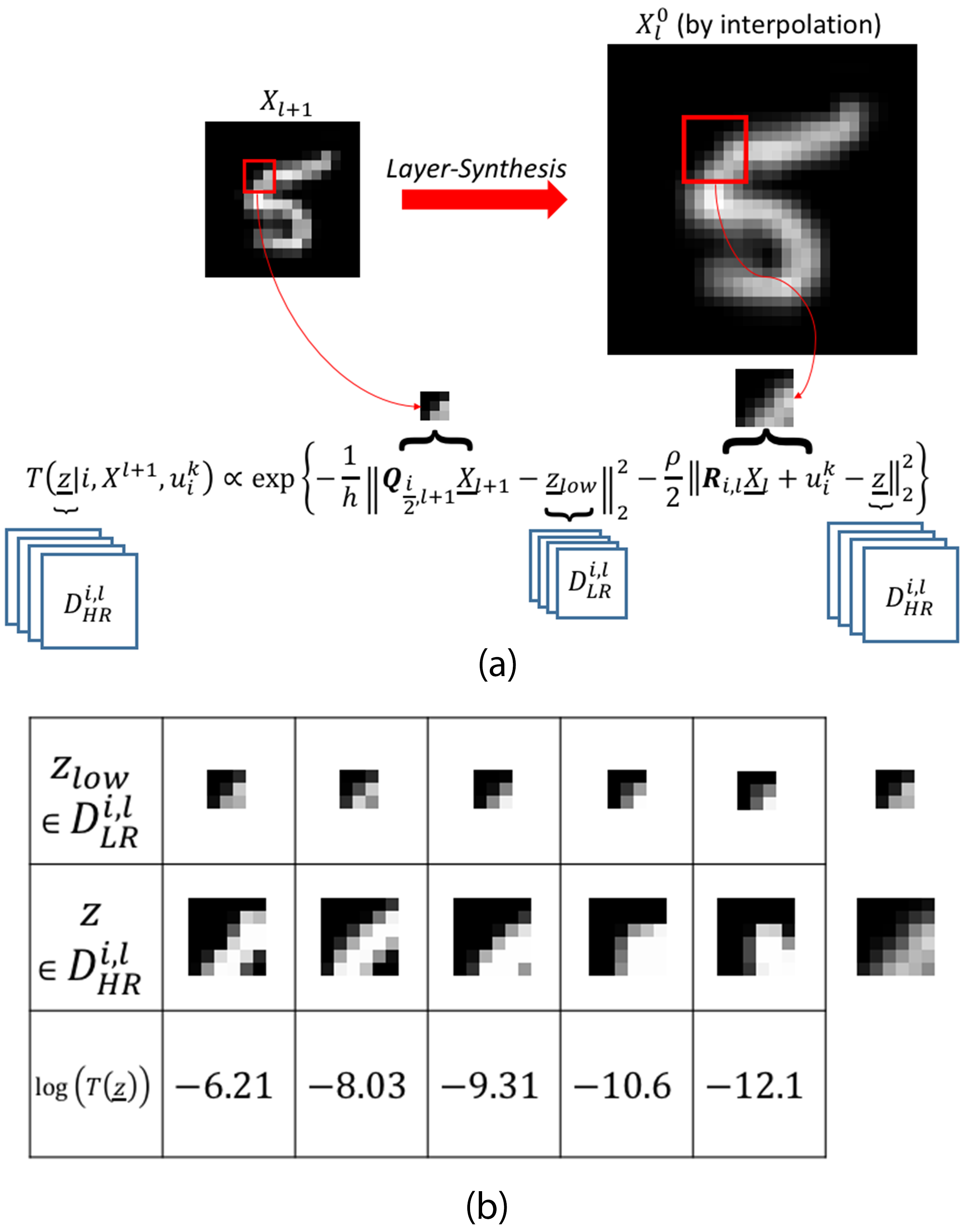}
    \caption{\textbf{(a)} The computation of the posterior distribution in Equation (\ref{eq:posterior example based}), from which a patch will be drawn for the refinement of $X_l$. \textbf{(b)} Example candidates of sampling from the distribution shown in Figure (a). Rightmost are the LR and HR measurements in patch-matching. Notice that the HR candidates are sharper than the current HR measurement, therefore improving the visual quality of the HR image at location $i$. In addition, as can be seen, the value of $\log(T(\underline{z}))$ is correlated with the relevance of the match.}
    \label{fig:sampling process}
\end{figure}

Next, we describe the definition of the patch priors based on $D_{LR}^{i, l}$ and $D_{HR}^{i, l}$, and show how these are used in the sampling process leading to the randomized synthesis effect. Algorithm \ref{alg:synthesis multi-scale} generates an image by inferring a Gaussian pyramid in the reversed order $X_{L},\ldots,X_0$, from the seed $X_{L}$ to the full image $X_0$. In particular, to generate $X_l$ (larger image) from $X_{l+1}$ (smaller one), the following process is invoked:
$$
\underline{X}_l = \textit{Layer-Synthesis}(\underline{X}_{l+1}, \{P_{i,l}\}_{i}),
$$
where we design the example-based prior $P_{i,l}$ (using $D_{LR}^{i, l}$ and $D_{HR}^{i, l}$ as defined above) to prefer HR patches that fit well to the LR content in $\underline{X}_{l+1}$. Formally, $P_{i,l}$ is defined only on existing patches from $D_{HR}^{i,l}$, and the probability assigned to the HR patch $D_{HR}^{i,l}(j)$ is dictated by the proximity between its LR version $D_{LR}^{i,l}(j)$ and the LR patch found at location $\frac{i}{2}$ in $X_{l+1}$, in the following way: 
\begin{equation} \label{eq:local prior}
P_{i,l}(D_{HR}^{i,l}(j) | \underline{X}_{l+1}) = \frac{1}{G_{i,l}} \cdot \exp \left\{-\frac{1}{h}\norm{\mathbf{Q}_{\frac{i}{2},l+1}\underline{X}_{l+1} - D_{LR}^{i,l}(j)}_2^2\right\}, \quad j = 1,\ldots, N
\end{equation}
where $G_{i,l}$ is the partition function. Note that we do not sample patches directly from the above prior. Instead, we shall plug it into Equation (\ref{eq:patch posterior}) and draw from the resulting posterior distribution:
\begin{multline} \label{eq:posterior example based}
T(D_{HR}^{i,l}(j) | i, \underline{X}_l^k, \underline{u}_i^k) = \frac{1}{G_i} \cdot \exp \Bigg\{-\frac{1}{h}\norm{\mathbf{Q}_{\frac{i}{2},l+1}\underline{X}_{l+1} - D_{LR}^{i,l}(j)}_2^2 
\\ - \frac{\rho}{2}\norm{\mathbf{R}_i \underline{X}^k_l + \underline{u}_i^k - D_{HR}^{i,l}(j)}_2^2\Bigg\},
\end{multline}
where $\underline{X}_l^k$ is the estimation of the final synthesized image at layer $l$ (denoted $\underline{X}_l$) after iteration $k$. Now we simply draw one of the patches from $D_{HR}^{i,l}$ to be used in location $i$, where the probability of each patch is determined by the above posterior. In fact, both the LR and HR measurements are taken into account simultaneously in Equation (\ref{eq:posterior example based}), as illustrated in Figure \ref{fig:sampling process}. This is in fact a patch-matching process where the LR measurement is matched with $D_{LR}^{i,l}$, and the estimated HR content is matched with $D_{HR}^{i,l}$.

Notice that the positive parameter $h$ in Equation (\ref{eq:posterior example based}) controls the randomness of the prior: if $h \rightarrow 0$, then the HR version of the nearest neighbor of $\mathbf{Q}_{\frac{i}{2},l+1}\underline{X}_{l+1}$ from $D_{LR}^{i,l}$ will be chosen with probability $1$, while $h \rightarrow +\infty$ makes the prior uniform over all the patches in $D_{HR}^{i,l}$. Through the iterations of the \textit{Layer-Synthesis}, we can change the parameter $h$ from a large initial value towards a smaller one which is proportional to $\rho$ (see Equation (\ref{eq:posterior example based})), so as to first allow new information to emerge and then, in later iterations, to make the HR image $X_l^{k+1}$ comply with both the previous estimate $X_l^k$ and the LR version $X_{l+1}$.

In order to obtain more relevant patches and speed up the algorithm in practice, we sample from the patches corresponding to the nearest neighbors of $\mathbf{Q}_{\frac{i}{2},l+1}\underline{X}_{l+1}$ (instead of from the whole candidate set $D_{HR}^{i,l}$), according to the probability assigned by Equation (\ref{eq:posterior example based}). Various work is suggested for efficient nearest neighbor searching \cite{barnes2009patchmatch,ben2007gray,muja2009fast,barnes2010generalized,xiao2011fast,he2012computing,olonetsky2012treecann,barnes2015patchtable}, and we decide to use the FLANN library \cite{muja2009fast} for this purpose.

\subsection{Extensions} \label{subsec:extensions}
In what follows we present three extensions to the above synthesis scheme that improve the synthesis process.

\subsubsection{Cycle-Spinning}

\begin{figure}
    \centering
    \includegraphics[width=0.9\textwidth]{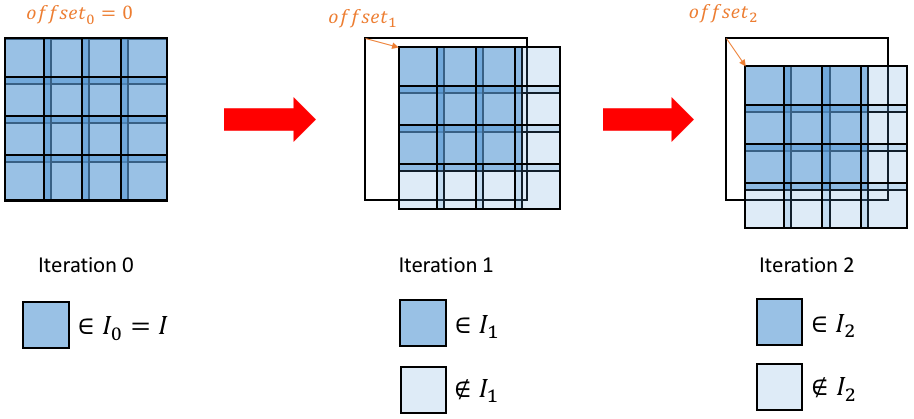}
    \caption{Illustration of cycle-spinning through iterations for \textit{Layer-Synthesis}. $I_k$ is the set of patches being refined in the $k$-iteration. The light-blue patches are not taken into account as they go outside the image support, for simplicity. }
    \label{fig:cycle spinning}
\end{figure}

We observe a trade-off between the visual quality and the richness of the outcome when choosing the overlap between patches. Concretely, fully overlapping patches (as in Algorithms \ref{alg:synthesis single} and \ref{alg:synthesis multi-scale}) avoid artifacts due to the averaging of many samples, while leading to blurry synthesized results which are similar to each other. On the other hand, with small overlaps we obtain sharp and diverse images, with the cost of artifacts that appear especially along the patch-borders. To bypass this trade-off, we suggest using the ``cycle-spinning" technique as follows. According to Figure \ref{fig:cycle spinning}, the leftmost image (corresponding to iteration $0$) is reconstructed from the patches at locations $i \in I_0 = I$ having small overlaps with each other (meaning that $I$ no longer contains all the patch locations). Then in the next iteration, as appears in the central part of Figure \ref{fig:cycle spinning}, we refine the previous estimate, this time by (randomly) restoring the patches at locations $i \in I$ shifted by $\text{offset}_1$ (the set of the shifted locations is denoted by $I_1$). Clearly, one can repeat this process with different offsets, thereby improving the visual quality due to the iterated refinement steps, while preserving the sharpness and richness of the result thanks to the small overlaps. A down side to this process is the need to hold local priors to all the candidate locations and invoke them in the restoration appropriately.

\subsubsection{Neighbor Search Window} 

As the training images may be slightly unaligned (e.g. digits and faces), in order to achieve more diverse results, the matching process should seek for similar patches in the database that are not only located in the same coordinates, but also in a small neighborhood around it. Formally, we propose to enlarge the dictionaries $D_{HR}^{i,l}$ and $D_{LR}^{i,l}$ built in Section \ref{build pair dictionary} to contain patches at location $i$ (as before), and also from the neighboring locations:
\begin{align*}
D_{HR}^{i, l} &= \bigcup_{i' \text{ neighbor of } i}\{\mathbf{R}_{i',l}\underline{V}_j^l\}_{j=1}^N, \\ D_{LR}^{i, l} &= \bigcup_{i' \text{ neighbor of } \frac{i}{2}}\{\mathbf{Q}_{i',l+1}\underline{V}_j^{l+1}\}_{j=1}^N,
\end{align*}
\noindent where $i'$ is a neighboring position of $i$ if the offset between $i$ and $i'$ is less than a given number of pixels in both $x$- and $y$-axes. This number is denoted as the $\text{neighbor-window-size}_l$ for layer $l$. Naturally, since digit and face images tend to be more ``aligned" in lower resolution, the window is increased together with the image size (see Tables \ref{table:digit parameters} and \ref{table:face parameters}).

In the extreme case, one could use all the patches in the images $X^l$ (larger) and $X^{l+1}$ (smaller) to form the dictionaries $D_{HR}^{i,l}$ and $D_{LR}^{i,l}$, respectively. This leads to a spatially invariant local model (in contrast to the spatially varying one that is defined above), which gives no influence to the locations of the patches. In coherent set of images as treated in this work, this approach is necessarily inferior.
%We provide several experiments with this option in Section \cmt{???}, in order to show the importance of locality in defining the patch-priors.

\subsubsection{Patch with Context} \label{patch context}
\begin{figure}
    \centering
    \includegraphics[width=0.8\textwidth]{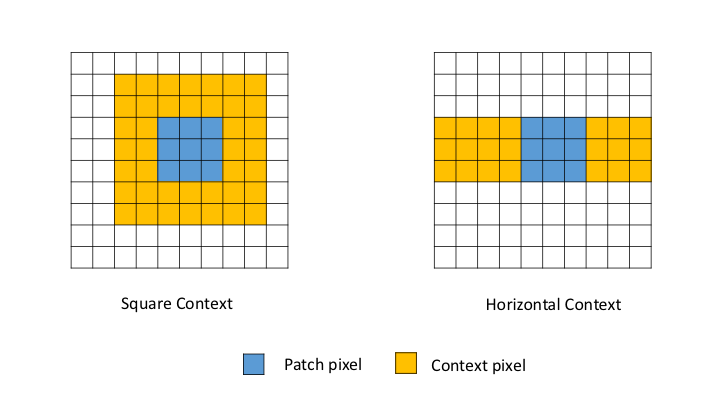}
    \caption{The square context and the horizontal context of an image patch.}
    \label{fig:patch context}
\end{figure}

When choosing the patch size to work with, the trade-off between image quality and originality arises: larger patch size leads to better visual quality, while limiting the richness of the generated images, and vice versa. The reason of this tendency is the limited size of the training set from which we can draw examples, along with the curse of dimensionality which implies that as patches get larger, relevant neighbors are becoming scarce.  To cope with this, the definition of distance between patches can be extended to take the surrounding area into account, as depicted in Figure \ref{fig:patch context}. Intuitively, as the patch associated with its context contains wider-range information, it will ensure that the high-level structure of the generated images will be more realistic, without sacrificing its originality. This approach is inspired by the recent work of Con-Patch \cite{romano2016patch}, which is shown to lead to better patch models. We stress here that this is not equivalent to working with bigger patches.

Later in the experiments, we use two kinds of contexts: the square context and the horizontal one, as illustrated in Figure \ref{fig:patch context}. The square context is used for both digit and face synthesis, while the horizontal context is used for faces to improve their symmetry.

\subsection{Comparison to \cite{kwatra2005texture}}

While the work reported in  \cite{kwatra2005texture} addressed a different task of texture-synthesis, it bares some similarities to the above described algorithm, due to the reliance on patch-matching, the operation in multi-scale and more. Here are the key differences between the two works:
\begin{itemize}
    \item This work deals with image synthesis in a more general way. As such, our treatment relies on local priors that are richer than the representation practiced in \cite{kwatra2005texture}.
    \item While in \cite{kwatra2005texture} randomness and originality of the result are not the prime goals, they are central in our scheme. In \cite{kwatra2005texture} only the initialization stage is random, while in our work the randomness is used in every patch-matching stage.
    \item The energy functional and minimization method in \cite{kwatra2005texture} are similar to the one posed by the EPLL (half-quadratic splitting), and therefore similar to ours as well. However, we introduce the ADMM, making the formulation more ``well-formed".
    \item Both methods use multi-scale pyramids, but \cite{kwatra2005texture} practices also a sweep over the patch sizes. We chose to avoid such feature in our algorithm, due to the anticipated problem of losing relevant neighbors in the consequent search, and the fear of getting large portions of trained images copied to the synthesized image.
    \item Both these works use small overlap between the patches and for the same reasons. our algorithm adds to this a cycle-spinning shift of the patch positions to avoid border artifacts. Perhaps \cite{kwatra2005texture} overcomes these artifacts due to the use of different patch-sizes.
    \item Last but not least, we accompany our algorithm with a framework for assessing the performance of whole image synthesis, as we will  outline in Section \ref{sec:assessment}.
\end{itemize}

\section{Synthesis Experiments}
\label{sec:synthesis experiments}

We now turn to present the results obtained by the described algorithm, synthesizing digits and face images.

\subsection{Synthesized Digits} \label{subsec:digits}

\begin{figure}
    \centering
    \includegraphics[width=0.8\textwidth]{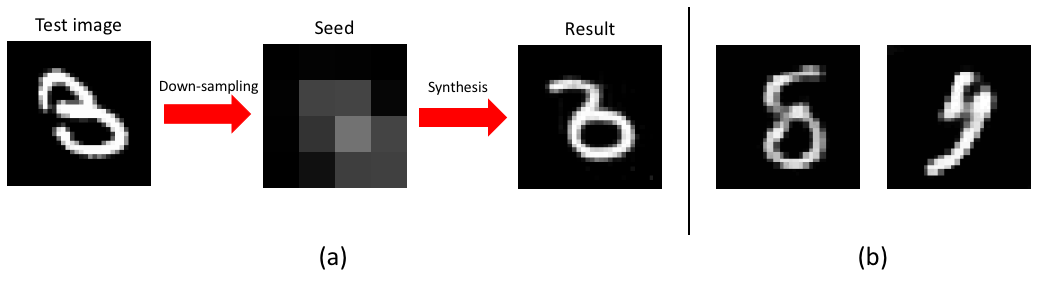}
    \caption{(a) Example of MNIST synthesis using a patch model built with all digits together. The result looks like the upper half of a digit ``3" merged with the lower half of a digit ``8". Notice that the LR seed image is indistinguishable from ``8". (b) Examples of failures in MNIST synthesis by DRAW\cite{gregor2015draw}, which also models all digits together.}
    \label{fig:all digit model example}
\end{figure}

\begin{figure}
    \centering
    \includegraphics[width=0.9\textwidth]{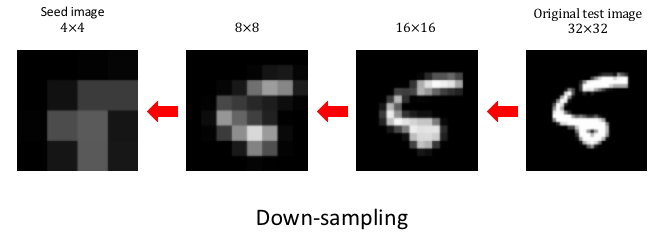}
    \caption{Down-sampling of one original MNIST test image to the size of a seed.}
    \label{fig:digit seed downsampling}
\end{figure}

Following the previous work in handwritten digit synthesis\cite{salakhutdinov2008quantitative,uria2014deep,raiko2014iterative,salakhutdinov2009deep,murray2009evaluating,rezende2014stochastic,salimans2015markov,gregor2014deep,gregor2015draw,radford2015unsupervised,van2016pixel,kingma2016improving,hinton2006fast,goodfellow2014generative,kim2016deep,germain2015made}, we use the MNIST \cite{lecun1998gradient} dataset in our experiments, which includes $60000$ training images and $10000$ test ones of size $28 \times 28$ pixels, padded to $32 \times 32$ before synthesis. The test images are downsampled to $4 \times 4$, which serve as the seed images (see Figure \ref{fig:digit seed downsampling}). It is worth noting that we choose to synthesize each type of digit using only the training images of the same digit (with the very same set of parameters), while existing work commonly builds the synthesis model with all kinds of digits together. Since the classification of MNIST digits is a very well studied problem for which highly accurate classifiers are available, we consider the classification as a simple preprocessing step. When we ignore the class-specific construction of the example set and use all the different digits together, images of ``non-digit" might be produced as shown in Figure \ref{fig:all digit model example}. A similar behavior, presented in the same figure, is observed when generating images by DRAW \cite{gregor2015draw}.\footnote{We used an unofficial implementation of DRAW, given in \cite{jang2016tensorflowdraw}.}

The synthesis consists of a pyramid of $4$ layers: $4 \times 4 \rightarrow 8 \times 8 \rightarrow 16 \times 16 \rightarrow 32 \times 32$. At each layer, our local example based patch prior (see Section \ref{subsec:local nn patch prior}) is used, along with the proposed extensions (as detailed in Section \ref{subsec:extensions}). The parameters (see Table \ref{table:digit parameters}) are chosen to achieve a good balance between the visual quality and the richness of the generated images.

\renewcommand{\arraystretch}{1.5}
\begin{table}[]
\centering
\caption{Parameters for digit synthesis}
\label{table:digit parameters}
\begin{tabular}{|c|l|}
\hline
parameters & values \\ \hline
$L$ -- pyramid depth & $\begin{array}{l} 3 \end{array}$ \\ \hline
image sizes & $\begin{array}{l} 4 \times 4 \rightarrow 8 \times 8 \rightarrow 16 \times 16 \rightarrow 32 \times 32 \end{array}$  \\ \hline
$n$ -- patch size & $\begin{array}{l} \end{array}$ $6$ for all layers \\ \hline
$\{\text{overlap}_l\}_{l = 0}^{L-1}$ (high-res) & $\begin{array}{l} \{2,2,2\} \end{array}$ \\ \hline
$\{\mathbf{H}_l\}_{l=0}^{L-1}$ -- downsampling & $\begin{array}{l}\end{array}$  $3\times 3$ Gaussian convolution ($\sigma = 1$) + $1/2$ decimation \\ \hline
$\{K_l\}_{l = 0}^{L-1}$ -- number of iterations & $\begin{array}{l} \{4,4,2\} \end{array}$  \\ \hline
$\{\text{neighbor-window-size}_l\}_{l = 0}^{L-1}$ & $\begin{array}{l} \{2,1,0\} \end{array}$ \\ \hline
$\begin{array}{c}\{\{h_{k,l}\}_{k=0}^{K_l - 1}\}_{l=0}^{L-1} \\ \text{(Equation (\ref{eq:local prior}))}\end{array}$ & $\begin{array}{l}\{\{2^{3},2^{4},2^{5},2^{6}\},\\ \{2^{3},2^{4},2^{5},2^{6}\},\\ \{100,100\} \}\end{array}$ \\ \hline
$\begin{array}{c}\{\{\lambda_{k,l}\}_{k=0}^{K_l - 1}\}_{l=0}^{L-1} \\ \text{(Equation (\ref{eq:epll admm}))}\end{array}$ & $\begin{array}{l}\{\{2^{-4},2^{-4},2^{-4},2^{-4}\},\\ \{2^{-4},2^{-4},2^{-4},2^{-4}\},\\ \{0,0.1\} \}\end{array}$ \\ \hline
$\begin{array}{c} \{\{\rho_{k,l}\}_{k=0}^{K_l - 1}\}_{l=0}^{L-1} \\ \text{(Equation (\ref{eq:epll admm}))}\end{array}$ & $\begin{array}{l}\{\{2^{-10},2^{-5},2^{0},2^{5}\},\\ \{2^{-10},2^{-6.67},2^{-3.33},2^{0}\},\\ \{0.01,100\} \}\end{array}$ \\ \hline
$\begin{array}{c} \{\{\text{offset}_{k,l}\}_{k=0}^{K_l - 1}\}_{l=0}^{L-1} \\ \text{(cycle-spinning)}\end{array}$ & $\begin{array}{l}\{\{[0,0],[1,0],[0,1],[1,1]\},\\ \{[0,0],[1,0],[0,1],[1,1]\},\\ \{[0,0],[0,0]\}\tablefootnote{$\{[0,0],[0,0]\}$ means that two regular ADMM iterations without cycle-spinning are performed here.} \}\end{array}$ \\ \hline
\end{tabular}
\end{table}

\begin{figure}
    \centering
    \includegraphics[width=0.7\textwidth]{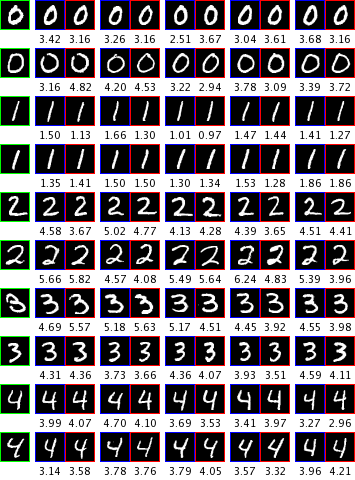}
    \caption{Example MNIST digits synthesis results (digits ``0" to ``4"). Green frame: original test image used to create the seed. Blue frame: different generation results with the seed in the same row. Red frame: the nearest neighbor of the synthesized image on its left from the training set. The distance between each generated image (blue frame) and its nearest neighbor (red frame on its right) is indicated below the blue frame, and the distance between this neighbor and its nearest neighbor from the training set is indicated under the red frame for comparison.}
    \label{fig:all digits 1}
\end{figure}

\begin{figure}
    \centering
    \includegraphics[width=0.7\textwidth]{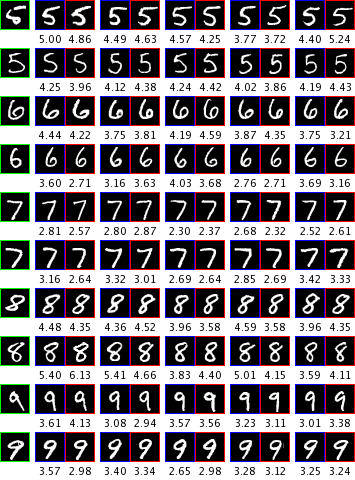}
    \caption{Example MNIST digits synthesis results (digits ``5" to ``9"). See the caption of Figure \ref{fig:all digits 1} for explanation.}
    \label{fig:all digits 2}
\end{figure}

\begin{figure}
    \centering
    \includegraphics[width=0.9\textwidth]{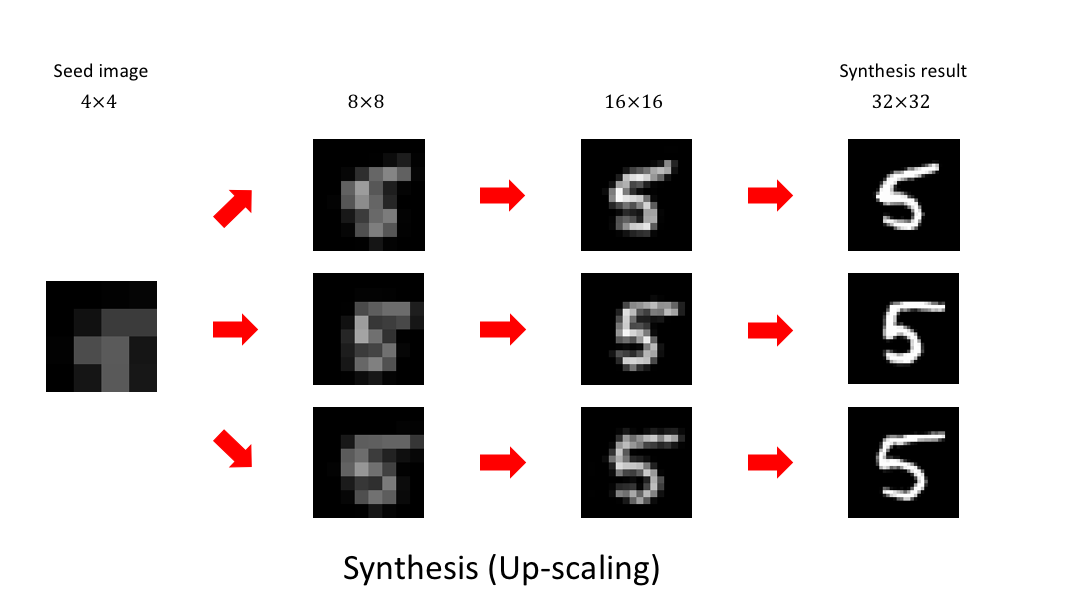}
    \caption{Different runs of synthesis of the digit ``5" using the same seed image.}
    \label{fig:digit gen branch}
\end{figure}

Figures \ref{fig:all digits 1} and \ref{fig:all digits 2} illustrates various generation results of all kinds of digits. As can be seen, the generated images have high visual quality. In addition, the results are different both from their nearest neighbor in the set of training images and the HR version of the seed image, i.e. our method successfully generated new images that are non-existing in the training set (good originality). Furthermore, as can be observed, we synthesize very diverse digits that originate from the same seed thanks to the randomness force, showing the effectiveness of our synthesis algorithm, and the good spread obtained over the training images. Given a seed image, a diagram of the synthesis process is depicted in Figure \ref{fig:digit gen branch}, showing how three different generations evolve throughout the layers.

% \begin{figure}
%     \centering
%     \includegraphics[width=0.8\textwidth]{gen_digits.png}
%     \caption{Example MNIST digits synthesis results. Green frame: original test image used to create the seed. Blue frame: different generation results with the seed in the same row. Red frame: the nearest neighbor of the synthesized image on its left from the training set.}
%     \label{fig:all digits}
% \end{figure}

\begin{figure}
    \centering
    \includegraphics[width=0.9\textwidth]{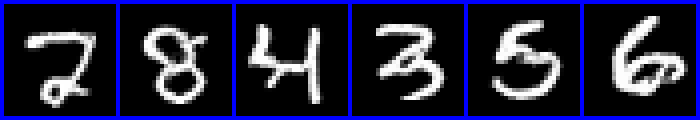}
    \caption{Example of failed MNIST synthesis using the proposed method.}
    \label{fig:our mnist failures}
\end{figure}

As observed in Figures \ref{fig:all digits 1} and \ref{fig:all digits 2}, there are no visible artifacts in the generated images. Nevertheless, we encountered several synthesis failures, which occur rarely, as shown in Figure \ref{fig:our mnist failures}. It is important to note that such failures can be discovered and explained by the LL measure, as described later in Sections \ref{subsec:ll} and \ref{subsec:assess mnist}.

\subsection{Faces} \label{subsec:faces}

% \cmt{Currently the results of non-aligned faces are not shown, as they should be further improved or left for future work.}

\renewcommand{\arraystretch}{1.5}
\begin{table}[]
\centering
\caption{Parameters for aligned face synthesis}
\label{table:face parameters}
\begin{tabular}{|c|l|}
\hline
parameters & values \\ \hline
$L$ -- pyramid depth & $\begin{array}{l} 4 \end{array}$ \\ \hline
image sizes & $\begin{array}{l} 8 \times 8 \rightarrow 16 \times 16 \rightarrow 32 \times 32 \rightarrow 64 \times 64 \rightarrow 128 \times 128 \end{array}$  \\ \hline
$n$ -- patch size & $\begin{array}{l} \end{array}$ $8$ for all layers \\ \hline
$\{\text{overlap}_l\}_{l = 0}^{L-1}$ (high-res) & $\begin{array}{l} \{2,2,2,4\} \end{array}$ \\ \hline
$\{\mathbf{H}_l\}_{l=0}^{L-1}$ -- downsampling & $\begin{array}{l}\end{array}$  $3\times 3$ Gaussian convolution ($\sigma = 1$) + $1/2$ decimation \\ \hline
$\{K_l\}_{l = 0}^{L-1}$ -- number of iterations & $\begin{array}{l} \{5,9,9,4\} \end{array}$  \\ \hline
$\{\text{neighbor-window-size}_l\}_{l = 0}^{L-1}$ & $\begin{array}{l} \{0,0,0,0\} \end{array}$ \\ \hline
% $\{\text{neighbor-window-size}_l\}_{l = 0}^{L-1}$ & $\begin{array}{l} \{0,0,0,0\} \text{ (aligned), } \{2,1,0,0\} \text{ (non-aligned)} \end{array}$ \\ \hline
$\begin{array}{c}\{\{h_{k,l}\}_{k=0}^{K_l - 1}\}_{l=0}^{L-1} \\ \text{(Equation (\ref{eq:local prior}))}\end{array}$ & $\begin{array}{l}\{\{2^{1},2^{-0.25},2^{-1.5},2^{-2.75},2^{-4}\},\\ \{2^{2},2^{1.25},2^{0.5},2^{-0.25},2^{-1},2^{-1.75},2^{-2.5},2^{-3.25},2^{-4}\},\\ \{2^{0},2^{-0.25},2^{-0.5},2^{-0.75},2^{-1},2^{-1.25},2^{-1.5},2^{-1.75},2^{-2}\}, \\ \{10,10,10,10\} \}\end{array}$ \\ \hline
$\begin{array}{c}\{\{\lambda_{k,l}\}_{k=0}^{K_l - 1}\}_{l=0}^{L-1} \\ \text{(Equation (\ref{eq:epll admm}))}\end{array}$ & $\begin{array}{l}\{\{2^{-3},2^{-2.25},2^{-1.5},2^{-0.75},2^{0}\},\\ \{2^{-3},2^{-2.625},2^{-2.25},2^{-1.875},2^{-1.5},2^{-1.125},2^{-0.75},2^{-0.375},2^{0}\},\\ \{2^{-3},2^{-2.625},2^{-2.25},2^{-1.875},2^{-1.5},2^{-1.125},2^{-0.75},2^{-0.375},2^{0}\}, \\ \{0.001,0.001,0.001,0.001\} \}\end{array}$ \\ \hline
$\begin{array}{c} \{\{\rho_{k,l}\}_{k=0}^{K_l - 1}\}_{l=0}^{L-1} \\ \text{(Equation (\ref{eq:epll admm}))}\end{array}$ & $\begin{array}{l}\{\{2^{-3},2^{-1.25},2^{0.5},2^{2.25},2^{4}\},\\ \{2^{-3},2^{-2.125},2^{-1.25},2^{-0.375},2^{0.5},2^{1.375},2^{2.25},2^{3.125},2^{4}\},\\ \{2^{-3},2^{-2.375},2^{-1.75},2^{-1.125},2^{-0.5},2^{0.125},2^{0.75},2^{1.375},2^{2}\}, \\ \{0.001,0.001,0.001,0.001\} \}\end{array}$ \\ \hline
$\begin{array}{c} \{\{\text{offset}_{k,l}\}_{k=0}^{K_l - 1}\}_{l=0}^{L-1} \\ \text{(cycle-spinning)}\end{array}$ & $\begin{array}{l}\{ \{[0,0],[2,0],[1,1],[0,2],[2,2]\},\\ \{[0,0],[1,2],[2,0],[0,1],[2,2],[1,0],[0,2],[2,1],[1,1]\},\\ \{[0,0],[1,2],[2,0],[0,1],[2,2],[1,0],[0,2],[2,1],[1,1]\},\\ \{[0,0],[1,0],[0,1],[1,1]\} \}\end{array}$ \\ \hline
\end{tabular}
\end{table}

Compared to digits, images of human faces are much more challenging to model and synthesize, as they contain richer details and very long range structures (e.g. the two ears should be consistent). Furthermore, humans are extremely sensitive to small unnaturalness of faces, making the generation task even more demanding.

In this experiment, we generate faces using a dataset of grayscale human faces from passport photos \cite{bryt2008compression,ram2014facial,elad2007low}. This dataset consists of 4500 example images and 500 test ones, all aligned by feature points (e.g. eyes and tip of nose). The alignment is done as in \cite{elad2007low}, where the feature points of each face are first located automatically, and then moved to the aligned locations by warping the triangles formed by these points using affine transformations. All the images (of size $221 \times 179$) are cropped to $179 \times 179$ and then resized to $128 \times 128$. The synthesis consists of a pyramid of $5$ layers: $8 \times 8 \text{ (seed) } \rightarrow 16 \times 16 \rightarrow 32 \times 32 \rightarrow 64 \times 64 \rightarrow 128 \times 128$. We use the very same algorithm as in the digit synthesis, but with different parameters (see Table \ref{table:face parameters}) due to the disparity between these two classes of images. % In addition, as a way to increase the quality and richness of the matches, we set a larger search window in the non-aligned faces compared to the aligned ones.

\begin{figure}
    \centering
    \includegraphics[height=0.7\textwidth,angle=-90,origin=c]{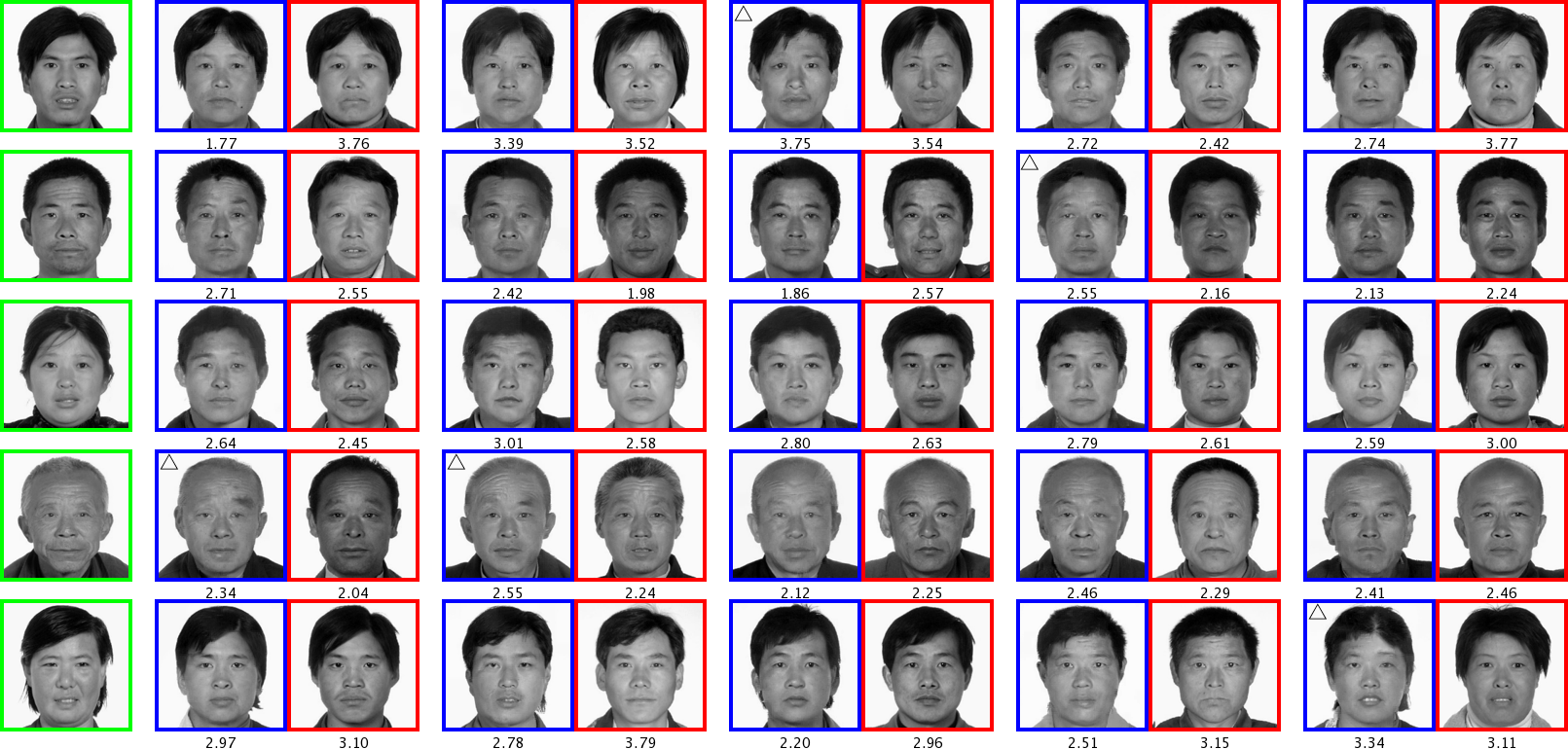}
    \caption{Example results of aligned faces synthesis. Refer to the caption of Figure \ref{fig:all digits 1} for the explanation of the color frames and the numbers. The distance used for searching the nearest neighbor takes into consideration only the central pixels, as they are most informative part in the face, while the boundary part (e.g. the hair and the clothes) has a lot of variation. The symbol $\triangle$ marks the faces that are slightly asymmetric.}
    \label{fig:face examples}
\end{figure}

\begin{figure}
    \centering
    \includegraphics[width=1.0\textwidth]{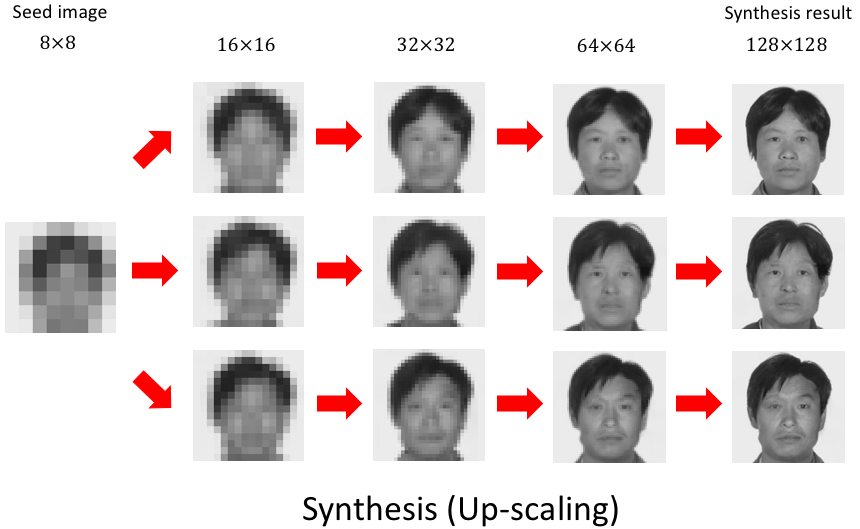}
    \caption{Different runs of aligned face synthesis using the same seed image.}
    \label{fig:face layers}
\end{figure}

Example of generated faces are depicted in Figure \ref{fig:face examples}. Visually, these synthesized images are realistic, containing fine details, and do not have severe artifacts. Furthermore, the ability of the proposed method to generate different faces from the same seed image is demonstrated in Figure \ref{fig:face layers}. This serves as an indication to effective randomness, leading to both rich and high quality results. As can be seen, despite the local nature of the proposed method, we obtain pleasant and realistic results. The symmetric property that is unique to faces is achieved by leveraging the horizontal context. As demonstrated, the eyes and ears in the generated images are aligned and have similar shape per instance. Minor asymmetry in some generated faces is still observed, marked by the symbol $\triangle$ in Figure \ref{fig:face examples}.

% The results of generation using a seed image from a  woman face with long hair may lead to man-like faces with long hair (Figure \ref{fig:face examples}, last row). In Figure \ref{fig:face layers} we see that the small facial details that distinguish men and women only appear at the two layers of highest resolutions, and most of them are too far from each other to be covered by the same patch (although we can increase the patch size, it may harm the originality of variability of the results). This reveals a limitation of our multi-scale local approach: it cannot handle long-range high resolution correlations very well.

\begin{figure}
    \centering
    \includegraphics[width=0.9\textwidth]{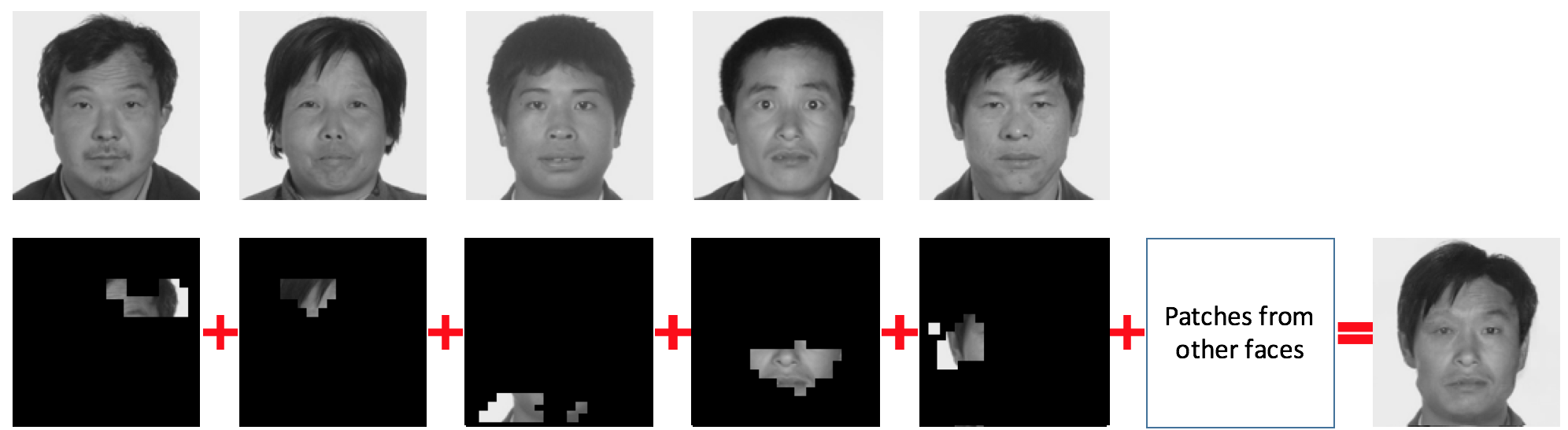}
    \caption{The result of one aligned face generation, shown as the stitching of patches from different images. First row: The first five training images contributing the most patches in the generated image. Second row: The patches these five images provide, and how they sum up to the synthesized image (with other patches). Notice that the new face is visually very different from the five faces on the left.}
    \label{fig:stitching}
\end{figure}

Differently from digit synthesis, in the case of faces, patches from the same training image tend to appear together in the generated result, forming coherent parts of a face (e.g. eye, nose; see Figure \ref{fig:stitching} for example). This behavior is similar to irregular patch stitching \cite{kwatra2003graphcut,efros2001image} in texture synthesis. This phenomenon leads to appealing results, but may indicate that more training examples are necessary, as we do not have enough freedom in combining the patches. On the down side, the patch stitching might deteriorate the originality of the generated images. However, hereafter in Section \ref{subsec:assess aligned faces}, we will show that the proposed algorithm does not suffer from this limitation.

% Finally, figure \ref{fig:nonaligned face examples} shows several examples of non-aligned face generation. While most of the synthesized faces are realistic as in the scenario of the aligned faces, some are less pleasant (e.g. first generation in the third row), and others contain severe failures in structure, e.g. deformed eyes (last generation in third row), missing mouth (first generation in last row), very asymmetric face (third generation in last row). In these failure cases, the generate images are easily distinguishable from the real ones. A possible source of the failures is the simplified patch matching step in the drawing process that does not combine mid-high level features. However, in order to show the ability of the proposed approach to handle a complete different type of images (faces and digits) we choose to avoid this possible improvement. As a promising future direction, we believe that stronger patch models for faces with semantic information of face structure may be necessary to handle the case of non-aligned faces.

% \begin{figure}
%     \centering
%     \includegraphics[width=0.9\textwidth]{nonaligned_gen_faces.png}
%     \caption{Example results of non-aligned faces synthesis.}
%     \label{fig:nonaligned face examples}
% \end{figure}

\section{Assessment of Synthesis Performance}
\label{sec:assessment}

As we have seen in Section \ref{sec:synthesis experiments}, our synthesis algorithm generates visually appealing and realistic digits and human faces. However, to assess the performance of our synthesis machine in a complete fashion, it is necessary to test if the generated images and the training ones represent the same probability distribution. As mentioned in Section \ref{sec:introduction}, evaluating the LL for either the test images or the generated images is insufficient for this task, and we suggest a complete assessment framework combining the following three indispensable components: the LL (Section \ref{subsec:ll}), the originality (Section \ref{subsec:originality}), and the spread (Section \ref{subsec:spread}). Then, we apply this framework to our digits and faces synthesis results in Sections \ref{subsec:assess mnist} and  \ref{subsec:assess aligned faces}, showing the strength and effectiveness of the proposed synthesis algorithm.

When comparing the scores of different generation methods, we believe that the three measures have the following importance order:
$$
\text{LL} \rightarrow \text{originality} \rightarrow \text{spread}.
$$
In words, if the LL of the generated images is exceedingly low, then they are of low visual quality or even meaningless, therefore the synthesis does not produce ``valid" images, and there is no point to evaluate the originality and spread. Similarly, if the generated images are too similar to the training ones (low originality), then the synthesis power is not effective, failing to fulfil one of the principal objectives of the process. In this case, a good spread score is misleading without high originality, as the generation results may be mostly replicated from the training set.

% In fact, the synthesis problem can be considered as finding local minima of the highly non-convex probability distribution function; and having a good distribution function does not guarantee one to find many different local minima of the function, hence the method of minimization does matter (which is also our main focus in this paper), and the examination of the generated images is necessary for the performance assessment.

\subsection{Log-Likelihood}
\label{subsec:ll}

The goal of the LL measure is multiple: (1) to show that the synthesis algorithm minimizes the EPLL objective function (up to some extent due to the randomness force); (2) to ensure the generated images have good visual quality, and detect the failure cases; (3)  to have a performance value which is comparable to previous work, for completeness.

\begin{table}
\centering
\begin{tabular}{l c}
\hline
\textbf{Model} & \textbf{LL Test} \\
\hline
DBM 2hl\cite{salakhutdinov2009deep}: & $\approx -84.62$ \\
DBN 2hl\cite{murray2009evaluating}: & $\approx -84.55$ \\
NADE\cite{uria2014deep}: & $-88.33$ \\
EoNADE 2hl (128 orderings)\cite{uria2014deep}: & $-85.10$ \\
EoNADE-5 2hl (128 orderings)\cite{raiko2014iterative}: & $-84.68$ \\
DLGM\cite{rezende2014stochastic}: & $\approx -86.60$ \\
DLGM 8 leapfrog steps\cite{salimans2015markov}: & $\approx -85.51$ \\
DARN 1hl\cite{gregor2014deep}: & $\approx -84.13$ \\
MADE 2hl (32 masks)\cite{germain2015made}: & $-86.64$ \\
DRAW\cite{gregor2015draw}: & $\geq -80.97$ \\
Diagonal BiLSTM (1 layer, $h=32$)\cite{van2016pixel}: & $-80.75$ \\
Diagonal BiLSTM (7 layers, $h=16$)\cite{van2016pixel}: & $-79.20$ \\
\hline
%$\sigma$ by t-SNE & $-89.65$ \\
suggested measure & $-82.52$ \\
\hline
\end{tabular}
\caption{Average LL values of \emph{test} images of the MNIST dataset (the LL values of the \emph{generated} images are provided later in Sections \ref{subsec:assess mnist} and \ref{subsec:assess aligned faces}). Larger (close to zero) is better. Note that the other methods use the same LL function for all digits, whereas our measure uses different priors for each type of digit.}
\label{table:average ll test images}
\end{table}

We define the likelihood of an image $X$ to be similar to the EPLL objective function as introduced in Equation (\ref{eq:epll}), which we aim to minimize during the synthesis process. Formally, we start by defining the following:
\begin{equation} \label{eq:ll pixel}
LL_{pixel}(X) = \frac{1}{\Tr(\sum_{i \in I} \mathbf{R}_i^T \mathbf{R}_i)}\sum_{i \in I} \log P_i(\mathbf{R}_i \underline{X}),
\end{equation}
where $P_i$ is the patch prior used in synthesis, and $\sum_{i \in I} \mathbf{R}_i^T \mathbf{R}_i$ is a diagonal matrix that counts for each pixel the number of different estimates emerging from the overlapping patches in $I$. As such, the term $1 / \Tr(\sum_{i \in I} \mathbf{R}_i^T \mathbf{R}_i)$ translates the sum over the LL of the patches $\mathbf{R}_i\underline{X}$ to the expected LL per pixel, denoted by $LL_{pixel}(X)$. Thus, the estimate of the LL of the whole image is given by
\begin{equation} \label{eq:ll image}
LL(X) = \abs{\underline{X}}\cdot LL_{pixel}(X),
\end{equation}
where $\abs{\underline{X}}$ denotes the number of pixels in $X$. Notice that this formulation is based on the EPLL assumption that all the patches $\mathbf{R}_i\underline{X}$ in the image $X$ are independent (even though they may be overlapping). As such, we sum their LL without conditioning.

While we aim at evaluating the LL of patches $\mathbf{R}_i\underline{X}$ that might not exist in the example patch database (as in Equation (\ref{eq:ll pixel})), our discrete priors are defined only on the existing patches in a dictionary (see Equation (\ref{eq:local prior})). Therefore, we suggest a continuous variant of the proposed discrete priors, formulated as a Parzen window\cite{parzen1962estimation} with Gaussian kernel:
\begin{equation} \label{eq:patch ll}
P_i (\underline{x}) = \frac{1}{\abs{D_{HR}^{i, 0}}} \sum_{\underline{y}_j \in D_{HR}^{i, 0}} \frac{1}{(2\pi \sigma^2)^{n^2/2}} \exp\left\{\frac{-\norm{\underline{x} -\underline{y}_j}^2}{2\sigma^2}\right\}
\end{equation}
where $D_{HR}^{i, 0}$ is the set of HR example patches, $\sigma^2$ stands for the window width, and $n^2$ denotes the number of pixels in a patch. Notice that we assume the covariance matrix of each Gaussian has full rank (i.e. equal to $\sigma^2 I$). Nevertheless, this full rank assumption is not valid in general, especially when considering background patches in both digits and faces. Specifically, all the patches in the background of the digits are totally flat, i.e. the term $\norm{\underline{x} -\underline{y}_j}^2$ is zero for all $j$ and for the totally flat $x$, leading to $P_i (\underline{x}) = 1 / (2\pi \sigma^2)^{n^2/2}$. Clearly, in this special case we expect to obtain $P_i (\underline{x}) = 1$. This is a direct consequence of the full-rank assumption, which is invalid for all $\sigma$ values different than $ 1 / \sqrt{2\pi}$. To cope with this singularity, one can suggest to estimate the rank of each covariance matrix (in the example above the rank should be $0$), however this approach raises various other difficulties. Another possible solution is to simply choose $\sigma = 1 / \sqrt{2\pi}$ so that the actual rank of the local covariance has no importance since $2\pi \sigma^2 = 1$. As a simple justification, we can view this $\sigma$ as representing a Gaussian noise level. In fact, $\sigma = 1 / \sqrt{2\pi}$ corresponds to a noise level of $17 / 255$ per pixel for $6 \times 6$ patches (digit) and $13 / 255$ for $8 \times 8$ patches (face), which is reasonable. Later in Appendix \ref{appendix:sigma} we provide more justifications for this value of $\sigma$.

Table \ref{table:average ll test images} provides the average LL value computed using Equations (\ref{eq:ll pixel}), (\ref{eq:ll image}) and (\ref{eq:patch ll}) on the test images of MNIST dataset. This table also lists the LL measure obtained by previous works. As can be inferred, our LL value is close to the state-of-the-art, indicating that the MNIST images are modeled well by our example-based local priors. Surprisingly, despite the simplicity and traceability of our approach, it is comparable to previous work which tends to be more complex.

As a closing remark, we emphasize that the above LL measure is utilized to evaluate the quality of the \emph{generated} images (see Sections \ref{subsec:assess mnist} and \ref{subsec:assess aligned faces}), instead of the \emph{test} ones as done in Table \ref{table:average ll test images}, which merely serves to align our modeling of the LL with the previous work.

% The evaluation uses the parameters for synthesis whenever possible (e.g. the patch size is $n = 6$, the overlap between patches is $2$)

\subsection{Originality} \label{subsec:originality}
In this subsection we propose a score that measures the originality/novelty of the generated images, which, in particular, ensures that the generated images are not simple replications of the training set. Intuitively, we can measure the distance of a generated image $X$ to the training images used for synthesis, denoted by the set $X_T = \{X_i | i \in T\}$. One can conclude that if this distance is very small, the originality of the image $X$ is limited. Formally, this distance is defined by
$$
d_G(X, X_T) \equiv \min_{X_i \in X_T} \norm{X - X_i}_2.
$$
Consequently, the nearest neighbor is denoted by
$$
X_{NN} = \NN(X; X_T) \equiv \argmin_{X_i \in X_T} \norm{X - X_i}_2
$$
Clearly, if the distance is extremely small the generated image is not novel as it is exceedingly similar to an existing example. However, a natural question that arises is how large this distance should be so one can truly consider this image as a new one. The answer we suggest is to compare $d_G(X, X_T)$ with the following distance:
$$
d_T(X, X_T) \equiv \min_{X_i \in X_T \setminus \{X_{NN}\}} \norm{X_{NN} - X_i}_2,
$$
where $A\setminus B$ denotes the set subtraction operation, and $X_{NN} = \NN(X; X_T)$. The above measures how close the training image $X_{NN}$ is to its own nearest neighbor in the database $X_T$ (which is different from itself). If $d_G(X, X_T) \geq d_T(X, X_T)$ we conclude that the generated image $X$ is indeed novel. On the other hand, $d_G(X, X_T) \ll d_T(X, X_T)$ implies that the synthesis algorithm does not hallucinate new data. This brings us to the definition of originality measure, given by
$$
Originality(X; X_T) = \frac{d_G(X, X_T)}{d_T(X, X_T)}.
$$
Thus, to quantify the overall originality of the whole set of generated images $X_G = \{X_i | i \in G\}$, we simply average the originality of all the individual generated images $X_i$:
$$
\mathbf{Originality}(X_G; X_T) = \frac{1}{|X_G|} \sum_{i \in G} Originality(X_i; X_T).
$$

We also propose a visual illustration of this measure, in which the 2-dimensional points $(d_G(X_i, X_T), d_T(X_i, X_T))$ are plotted for each image $X_i \in X_G$ (refer to Figure \ref{fig:mnist originality visualization} as an example to such visualization that compares our generation to the results of DRAW).

\subsection{Spread} \label{subsec:spread}
Consider a scenario in which a synthesis algorithm produces one single image that has high LL and originality. In this case, the overall performance is poor due to the lack of richness and diversity of the results. Motivated by this, we suggest a measure of spread to ensure that the generated images are as ``spread out" as the training ones.

We suggest to assess the spread of the generated images mainly based on the t-SNE \cite{van2008visualizing} unsupervised non-linear embedding. The motivation is that t-SNE is widely used and produces state-of-the-art result in MNIST visualization, in which the different types of digits are visually separated into different clusters (while other popular embedding techniques such as Isomap \cite{tenenbaum2000global} and LLE \cite{roweis2000nonlinear} do not). Therefore, we believe that t-SNE is also able to reveal the considerable inconsistencies between the training images and the generated ones, if any.

Next, we review briefly the main principle of the t-SNE applied on an image set $X_D = \{X_i | i \in D\}$. In this technique, the probability of the image $X_j \in X_D$ being the neighbor of $X_i \in X_D$, denoted by $p_{j|i}$, is defined as:
\begin{equation} \label{eq:tsne weight}
p_{j|i} = \frac{\exp(-\norm{X_i - X_j}_2^2 / 2\sigma_i^2)}{\sum_{k\in D,k \neq i} \exp(-\norm{X_i - X_k}_2^2 / 2\sigma_i^2)}
\end{equation}
where $\sigma_i$ is chosen such that
$$
H(P_i) = -\sum_{j} p_{j|i} \log p_{j|i}
$$
has a given fixed value for all $i \in D$. In fact, $p_{j|i}$ defines a weighted neighborhood of the point $X_i$, i.e. any image $X_j \in X_D$ is neighbor of $X_i$ with weight $p_{j|i}$. Next, for the actual embedding task, denote by $y_i$ the 2-dimensional embedded point of $X_i$, and $q_{j|i}$ the neighborhood probability for the embedded points, similarly to $p_{j|i}$:
$$
q_{j|i} = \frac{\exp(-\norm{y_i - y_j}_2^2)}{\sum_{k\in D,k \neq i} \exp(-\norm{y_i - y_k}_2^2)}
$$
As such, the objective function to minimize with regard to (w.r.t) the embedded points $y_i$ is given by
$$
C = \sum_{i} KL(P_i || Q_i) = \sum_{i} \sum_{j} p_{j|i} \log(\frac{p_{j|i}}{q_{j|i}}).
$$
After minimizing $C$, the points $y_i$ represent the two most significant components of the geometry of the set $X_D$, which can be plotted on a 2-dimensional plan for visualization (see the embedding result in Figure \ref{fig:mnist tsne}, for example. Notice that the points are visually separated into 10 clusters, as there are 10 types of digits).

The success of t-SNE shows that $p_{j|i}$ preserves important information of the manifold of images, which should be beneficial for the measure of spread in the context of image synthesis. Denoting by $X_T = \{X_i | i \in T\}$ the set of training images and by $X_G = \{X_i | i \in G\}$ the set of generated ones\footnote{For a fair comparison between the two sets, we assume that they have about the same number of elements: $|X_T| \approx |X_G|$.}, the proposed spread measure around one training image $X_i \in X_T$ is defined to be the ratio between the density of the neighbors of $X_i$ from $X_G$ and its neighbors from $X_T$:
$$
Spread(i;X_T,X_G) = \log\left(\frac{\sum_{j \in G} p_{j|i} \norm{X_j - X_i}_2^2}{\sum_{j \in T} p_{j|i} \norm{X_j - X_i}_2^2}\right),
$$
where $p_{j|i}$ is defined as in Equation (\ref{eq:tsne weight}) by merging the training set with the set of generated images, i.e. $D = T \cup G$\footnote{We stress on the fact that this spread measure does not use the directly the embedding result of t-SNE $\{y_i\}_{i \in D}$. Instead, it is merely based on the pairwise distances of images and the probabilities $\{p_{j|i}\}_{i,j}$.}. The sum $\sum_{j \in G} p_{j|i} \norm{X_j - X_i}_2^2 + \sum_{j \in T} p_{j|i} \norm{X_j - X_i}_2^2$ can be seen as the trace of the covariance matrix of all the neighbors of $X_i$ from $X_G$ and $X_T$\footnote{the neighbor $X_j$ is weighted by $p_{j|i}$. The mean of the neighbors is assumed to be $X_i$.}. The term $\sum_{j \in G} p_{j|i} \norm{X_j - X_i}_2^2$ is the contribution of $X_G$ to this trace, and $\sum_{j \in T} p_{j|i} \norm{X_j - X_i}_2^2$ is the contribution of $X_T$ to it. Intuitively, as $|X_T| \approx |X_G|$, $Spread(i;X_T,X_G) \approx 0$ indicates that $X_i$ has about the same number of neighbors from $X_T$ and $X_G$ within the radius $\sigma_i$, so there is little bias in synthesis from the ``point of view" of $X_i$. If $Spread(i;X_T,X_G) \gg 0$, then $X_G$ has a much larger density of points around $X_i$ compared to $X_T$, and vice versa for the case $Spread(i;X_T,X_G) \ll 0$.

Having the spread measure for each training image defined, we can score the overall performance simply by the average of the the spread of all training images:
$$
\mathbf{Spread}(X_G; X_T) = \frac{1}{|X_T|} \sum_{i \in T} \abs{Spread(i;X_T,X_G)},
$$
which is expected to be as small as possible. %Also, we can visualize the spread score by plotting the training points at their 2D position embedded by t-SNE, where each point $X_i$ is assigned to the color corresponding to its individual spread score $Spread(i;X_T,X_G)$ in the color map.

Finally, to visualize how the synthesized images spread out over the training ones, we simply apply t-SNE on the training images $X_T$ and generated ones $X_G$ altogether, and plot the 2-dimensional embedding result, using different colors for images from $X_T$ and $X_G$. This way, one can directly observe if the elements of $X_T$ and $X_G$ overlap well in the embedded space.

\subsection{Assessment of MNIST Synthesis} \label{subsec:assess mnist}

In this section we apply the assessment framework described above to compare our generated MNIST digits to the ones of state-of-the-art DRAW\cite{gregor2015draw}\footnote{As DRAW is the only recent and leading work for which we found an online implementation \cite{jang2016tensorflowdraw}.}. Table \ref{table:mnist assessment scores} shows these scores for the synthesized- and the test-images as a reference performance.
% Notice that for the LL and the originality, we need to know the type of the generated digits to evaluate the score. However, in DRAW the type of the digits are unknown, and the cope with it, we simply use a very common convolutional neural classifier of MNIST digits, which has an error rate lower than $1\%$, seemingly good enough for the assessment task.

\begin{table}
\centering
\begin{tabular}{l c c c}
\hline
\textbf{Model} & \textbf{LL} & \textbf{Originality} & \textbf{Spread} \\
\hline
DRAW\cite{gregor2015draw} & -79.33 & 0.622 & \textbf{1.11} \\
suggested method & \textbf{-68.20} & \textbf{1.04} & 1.31 \\
\hline
test images & -82.52 & 1.09 & 0.911 \\
\hline
\end{tabular}
\caption{Scores of the assessment framework on \emph{generated} (first two rows) and \emph{test} (third row) MNIST digits. The scores of images of different digits are averaged together. Bold indicates the best performance in synthesis. Notice that the first two LL values are computed on \emph{generated} images, in contrast to the values shown in Table \ref{table:average ll test images}, which are evaluated on \emph{test} images.}
\label{table:mnist assessment scores}
\end{table}

\begin{figure}
    \centering
    \includegraphics[width=0.9\textwidth]{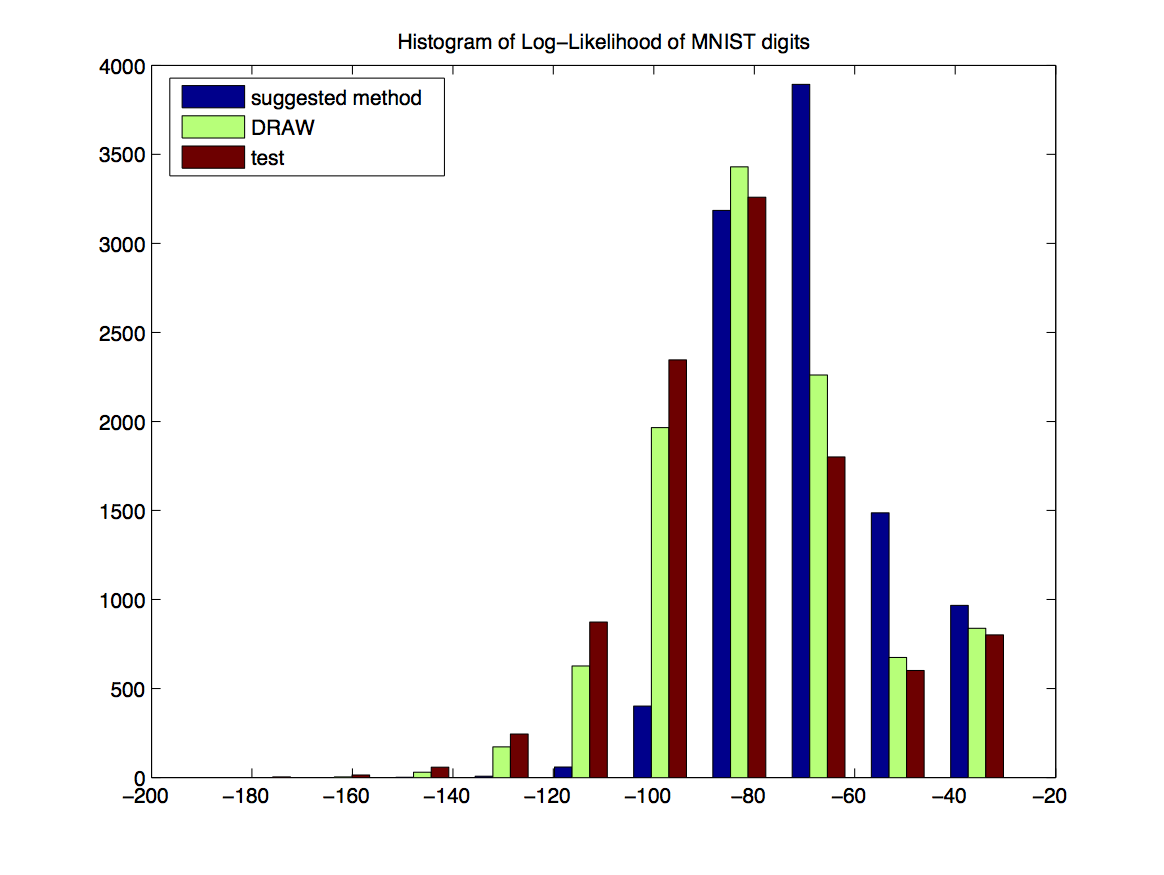}
    \caption{Histogram of the LL of the generated MNIST images and the test images.}
    \label{fig:mnist ll histogram}
\end{figure}

For the LL (the higher the better), our average value is much higher than DRAW and the one obtained on the test images, implying that the generated images are of high visual quality with regard to the training set. This is not surprising, as our method targets the minimization of the sum of the LL of the patches. Notice that the test images have smaller LL than both synthesis methods, for which a possible explanation is that patches of the test images might not exist in the training set, resulting in a lower patch LL. In Figure \ref{fig:mnist ll histogram} we plot the histogram of the LL values of the different images, showing that the distribution of LL of DRAW and our method are relatively close to that of the test images.

\begin{figure}
    \centering
    \includegraphics[width=0.9\textwidth]{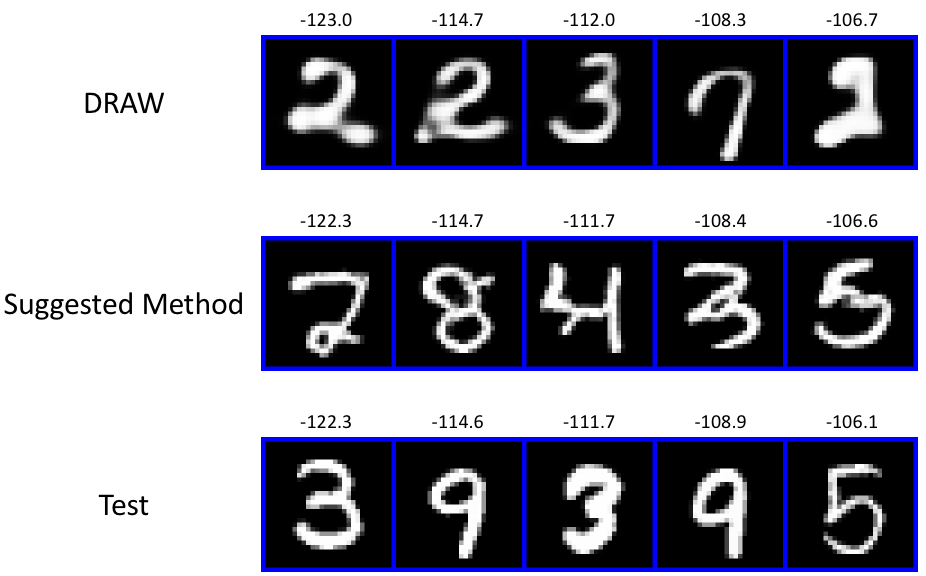}
    \caption{Examples of MNIST images with low LL values. The LL of each image is indicated above it. From up to down, The images are generated by DRAW, the proposed method and taken from the test set, respectively.}
    \label{fig:mnist low ll}
\end{figure}

Notice that low LL score indicates that the image is unique, however, it cannot distinguish whether this unique result is a failure or a truly novel and realistic generated image. This explains why the LL of the test images is lower than the generated ones. On the other hand, we empirically observe that low LL measure is correlated with badly \emph{generated} images, but not with bad \emph{training/test} ones, as depicted in Figure \ref{fig:mnist low ll}. 

% the generated images are expected to replicate the distribution of the training images. Our way to visually confirm this is the histogram of LL values (Figure \ref{fig:mnist ll histogram}): Although our images tend to have larger LL values than the test images, the histogram of our images is still reasonably similar to that of the test images. On the other hand, the histogram of DRAW is quite faithful to that of the test images.

\begin{figure}
    \centering
    \includegraphics[width=0.9\textwidth]{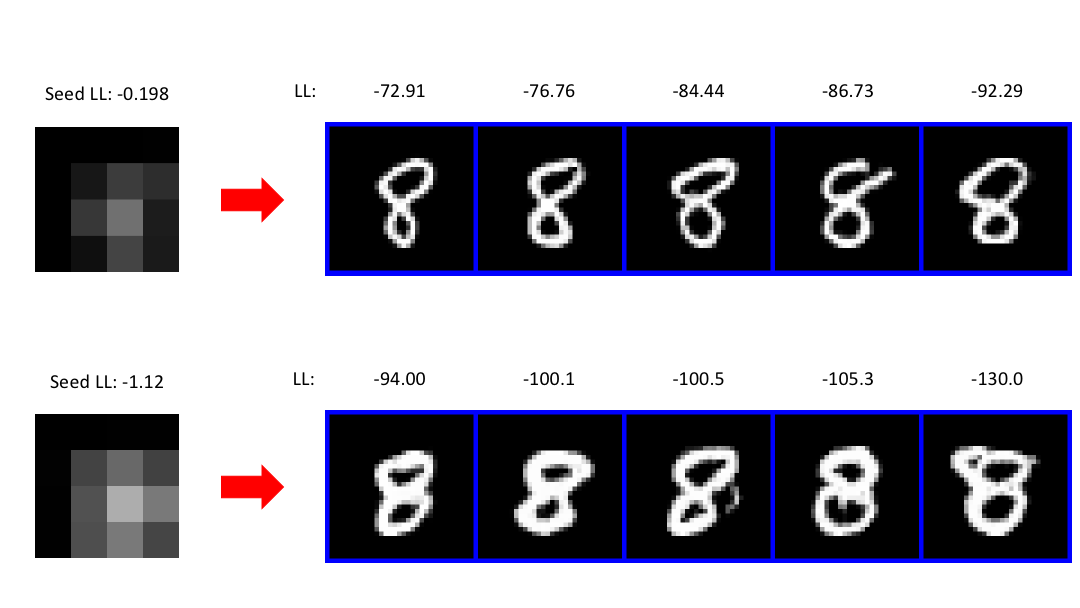}
    \caption{Two seeds of digit the ``8" with different LL and their generation results. The images are sorted by decreasing LL. The LL of a seed image is evaluated under the Parzen window defined with the seeds  created by downsampling all the training images. The average of LL of all the seeds of ``8" is -0.279.}
    \label{fig:mnist failure example}
\end{figure}

As such, we use the LL measure to detect and understand the failed generations. In fact, we observe that the failure is mostly related to the seeds: Digits of high quality are generated from most seeds, while some seeds lead more frequently to failures. To explain this difference, we can compute the LL of these two kinds of seeds\footnote{We model the probability of the $4\times 4$ seeds by the Parzen window defined by the seed images created from the whole training set, using Gaussian kernel.}, and compare their synthesis results. Figure \ref{fig:mnist failure example} shows two example seeds of the digit ``8" and different generation runs from them, with the LL of each image. The first seed has a high LL, and its results have good visual quality and contain no visible failures; On the other hand, the second seed has a much lower LL, leading to moderate failures. Intuitively, the LL of the seed indicates its uniqueness, and the difficulty to create likely images from it. Also notice that in the results of the second seed, the visual quality decreases with the LL (from left to right), showing that our LL definition is indeed meaningful.

\begin{figure}
    \centering
    \includegraphics[width=0.9\textwidth]{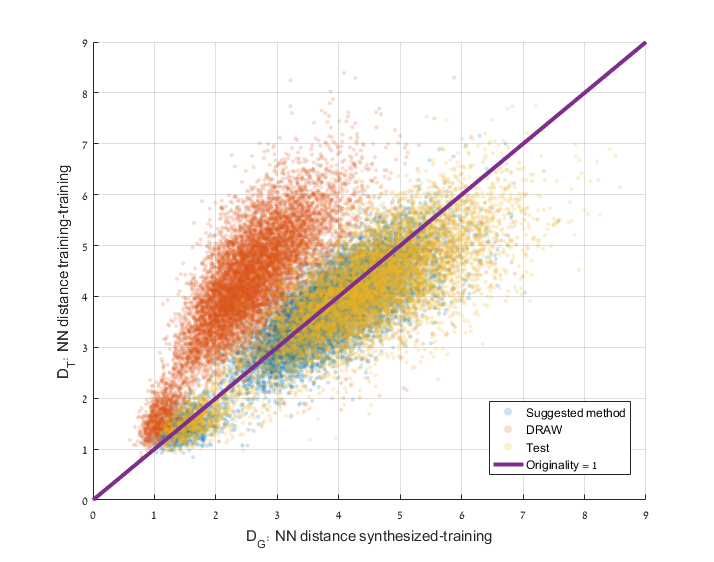}
    \caption{Visualization of the originality of the MNIST digits synthesized by the suggested method and DRAW, and the MNIST test images.}
    \label{fig:mnist originality visualization}
\end{figure}

As for the second assessment score -- the originality (higher is better) -- our generated images have an originality value close to $1$, implying that the proposed method synthesizes rich and novel images which are far from being replicated version of the training images. As a comparison, the images of DRAW have much smaller originality (equal to 0.622), as shown in Table \ref{table:mnist assessment scores}. Visually, Figure \ref{fig:mnist originality visualization} plots the points $(D_G, D_T)$ (as defined in Section \ref{subsec:originality}) for our results, the results of DRAW and the test images. The points corresponding to images with good originality are those close to the line $D_G = D_T$ or below it. As can be seen, our generated images have high originality as the test images do, which is greater than the originality of DRAW.

\begin{figure}
    \centering
    \includegraphics[width=1.0\textwidth]{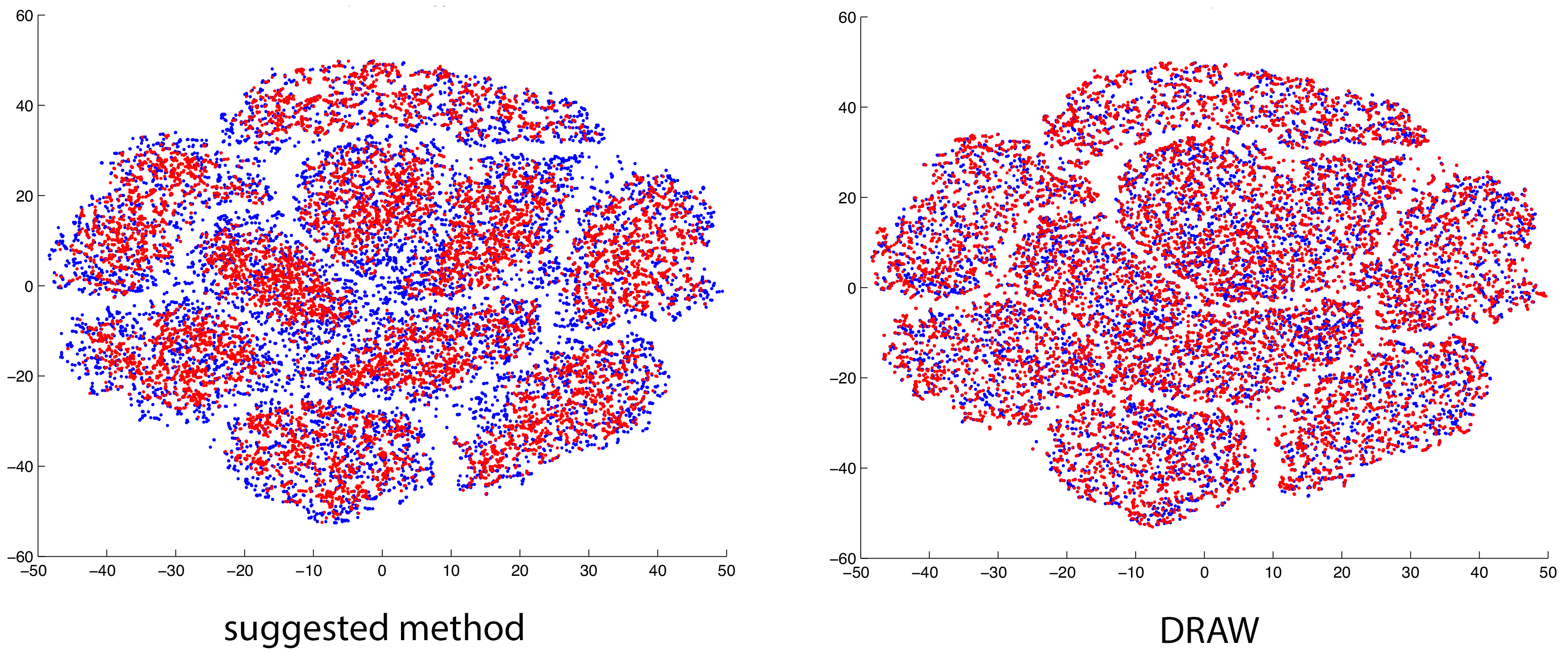}
    \caption{t-SNE visualization result of the MNIST digits generated by the suggested method (red points on the left) and DRAW (red points on the right), together with part of the training images (blue points). The $x$ and $y$-axes represent the two most significant geometry components found by t-SNE.}
    \label{fig:mnist tsne}
\end{figure}

\begin{figure}
    \centering
    \includegraphics[width=1.0\textwidth]{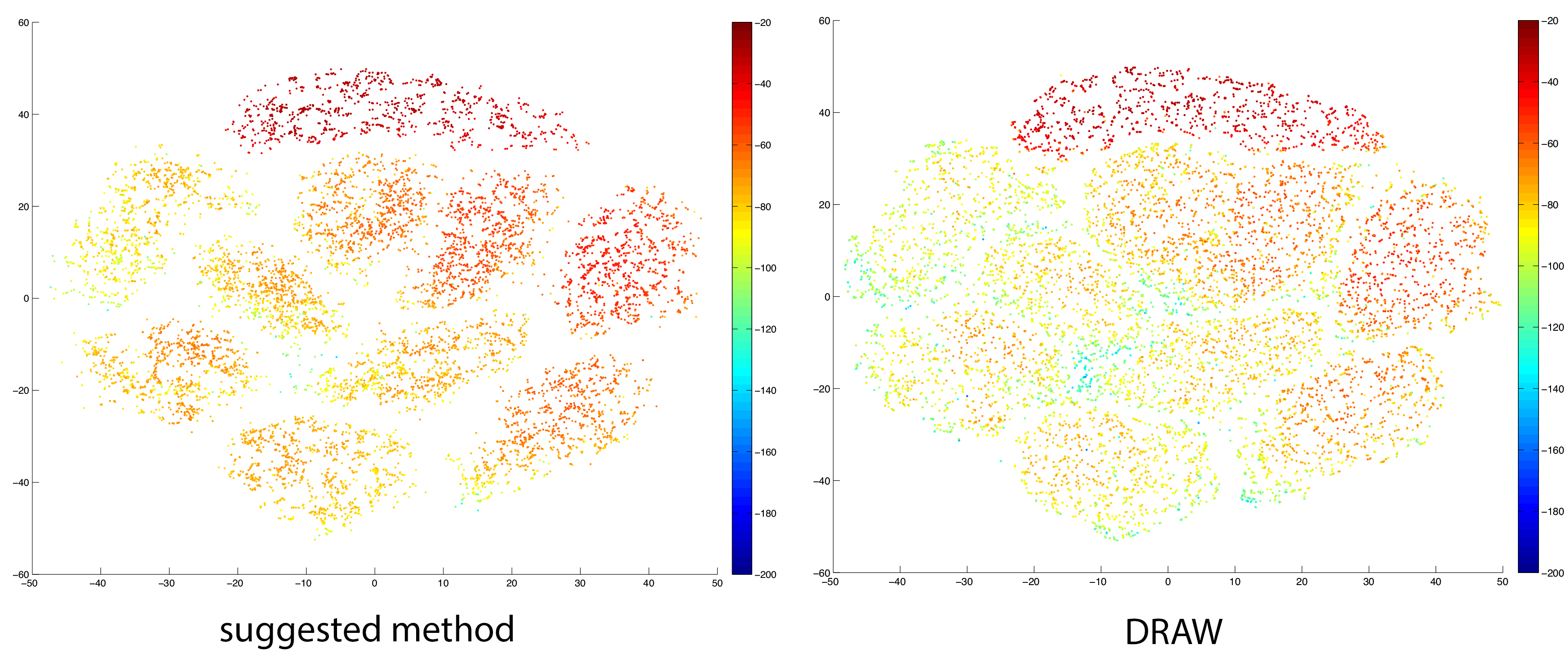}
    \caption{Visualization of the LL of the MNIST digits generated by the suggested method (left) and DRAW (right). The positions of the points are the same as the red points in Figure \ref{fig:mnist tsne}. The colors of the points correspond to the LL value of the images.}
    \label{fig:mnist tsne ll}
\end{figure}

\begin{figure}
    \centering
    \includegraphics[width=0.9\textwidth]{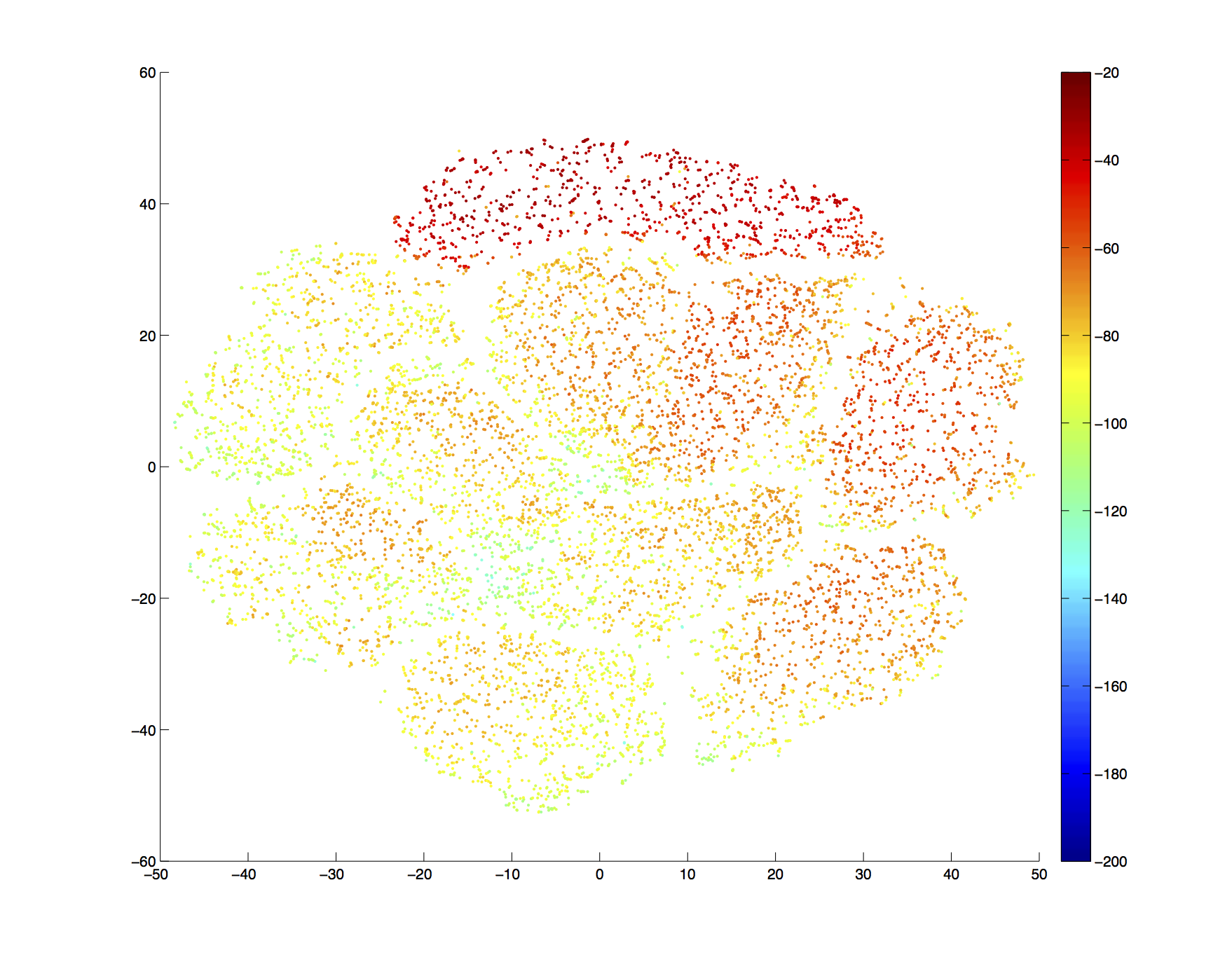}
    \caption{Visualization of the LL of part of the MNIST training images. The positions of the points are the same as the blue points in Figure \ref{fig:mnist tsne}. The colors of the points correspond to the LL value of the image.}
    \label{fig:mnist train tsne ll}
\end{figure}

Now we turn to the last measure, the spread. Following Table \ref{table:mnist assessment scores}, DRAW has a better spread score than our method, which is confirmed by the t-SNE visualizations in Figure \ref{fig:mnist tsne}. As can be observed, the points corresponding to the generated images of DRAW overlap well the points of the training images (right figure), while the points corresponding to our generated images are ``biased" towards the regions more populated by the points of the training images (left figure). The lower spread of our result can be explained by the LL measure. As shown in Figures \ref{fig:mnist tsne ll} and \ref{fig:mnist train tsne ll}, the LL value of the images in less populated regions have slightly lower LL values. Since our method maximizes the LL to some extent, the images are ``dragged" towards more populated regions during the synthesis process. On the other hand, as DRAW tends to replicate training images, its LL plot (Figure \ref{fig:mnist tsne ll}) is also visually quite similar to the LL plot of the training images (Figure \ref{fig:mnist train tsne ll}).

\begin{figure}
    \centering
    \includegraphics[width=1.0\textwidth]{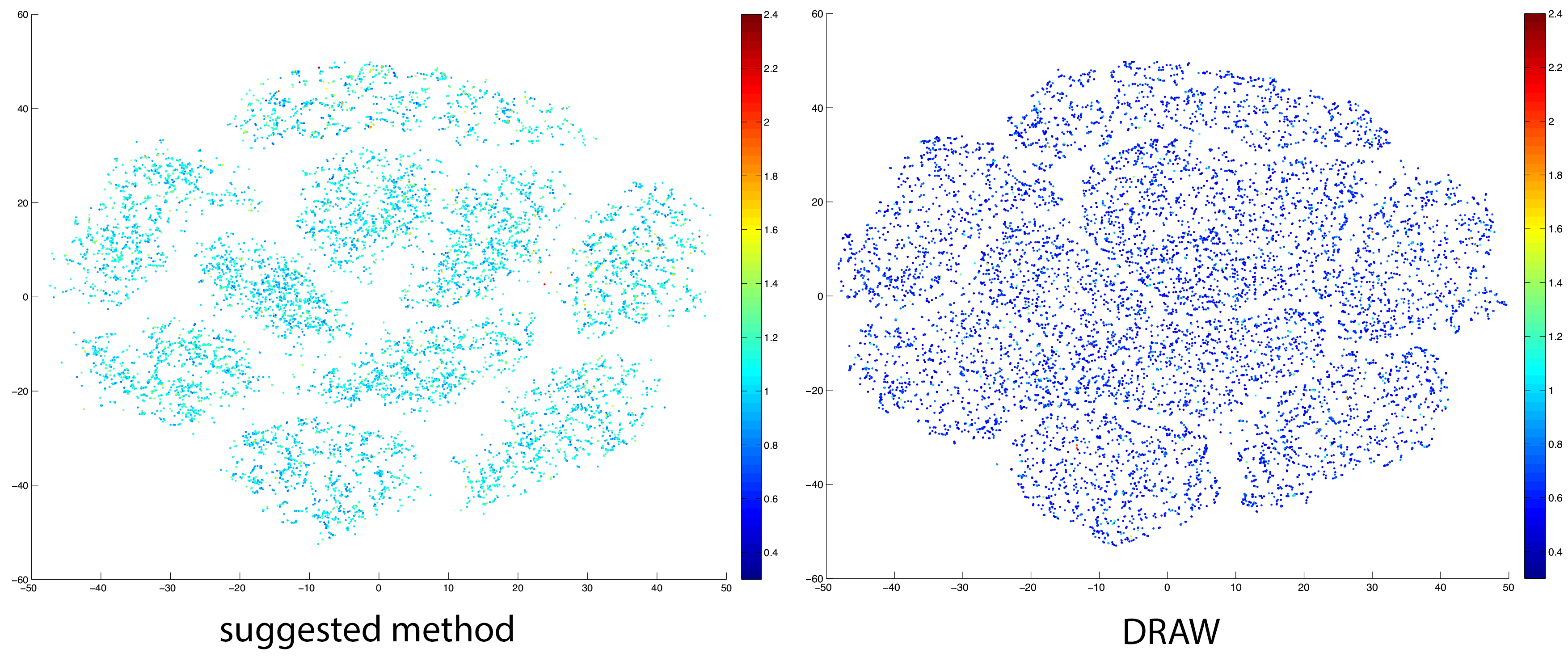}
    \caption{Visualization of the originality of the MNIST digits generated by the suggested method (left) and DRAW (right). The positions of the points are the same as the red points in Figure \ref{fig:mnist tsne}. The colors of the points correspond to the originality value of the images.}
    \label{fig:mnist tsne originality}
\end{figure}

In conclusion, our synthesis algorithm produces high quality MNIST images, outperforming DRAW in both LL and originality. Although DRAW has a better spread, it may be due to its tendency to replicate training images (as its originality score indicates).

\subsection{Assessment of Aligned Face Synthesis} \label{subsec:assess aligned faces}

In this subsection we apply the the proposed assessment to the aligned faces. As no previous work uses this face data set for synthesis, we present the suggested scores and visualizations for our generated images, and for 500 test images for comparison. A total amount of 1000 faces are generated using the 500 seeds (of size $8 \times 8$) that are created from the test images, resulting in the scores that are shown in Table \ref{table:aligned faces assessment scores}.

\begin{table}
\centering
\begin{tabular}{l c c c}
\hline
\textbf{Model} & \textbf{LL} & \textbf{Originality} & \textbf{Spread} \\
\hline
suggested method & -370.1 & 1.04 & 1.28 \\
\hline
500 test images & -427.0 & 1.13 & 1.79 \\
\hline
\end{tabular}
\caption{Scores of the assessment framework on \emph{generated} (first row) and \emph{test} (second row) aligned faces.}
\label{table:aligned faces assessment scores}
\end{table}

Similarly to digit synthesis, the LL of the generated faces are much higher than the values obtained for the test images, indicating high visual quality. In fact, due to high sensitivity of human eyes to faces (in contrast to artificial images such as digits), we choose the synthesis parameters so as to avoid visible artifacts for reasonable visual quality, while slightly limiting the randomness. For this reason, low LL values do not indicate failed generations.

\begin{figure}
    \centering
    \includegraphics[width=0.7\textwidth]{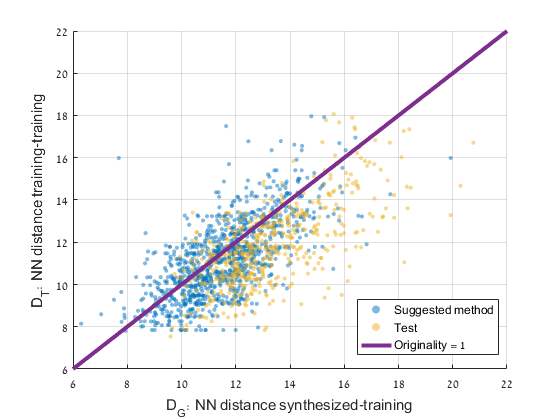}
    \caption{Visualization of the originality of the synthesized aligned faces, and faces from the test set.}
    \label{fig:aligned face originality}
\end{figure}

\begin{figure}
    \centering
    \includegraphics[width=0.9\textwidth]{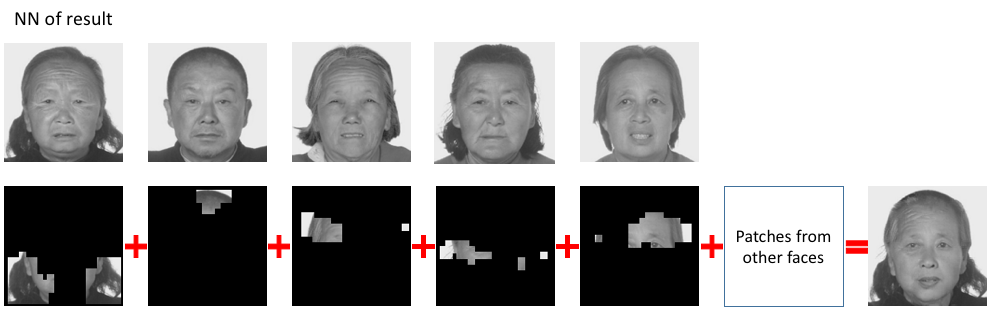}
    \caption{Example of generated aligned face with poor originality (0.481), and how it is formed by patches from different training images. The first training image contributes a lot of patches, and it is the nearest neighbor of the synthesis result.}
    \label{fig:aligned face originality failure example}
\end{figure}

The visualization of the originality is shown in Figure \ref{fig:aligned face originality}, illustrating that the generated images are not replications of the existing training images. Moreover, these are close to a very good originality that is equal to $1$ (also as shown in Table \ref{table:aligned faces assessment scores}), although slightly lower than the test images. Nevertheless, few generations may have poor originality due to the choice of too many patches from the same training image. Such an example is shown in Figure \ref{fig:aligned face originality failure example}, where the bottom part of the hair originates from the same image. This can be explained by the fact that the database is relatively small, and if unique patches (such as hair) are chosen, the algorithm will prefer to keep the consistency between the patches with the cost of reduced originality.

\begin{figure}
    \centering
    \includegraphics[width=0.9\textwidth]{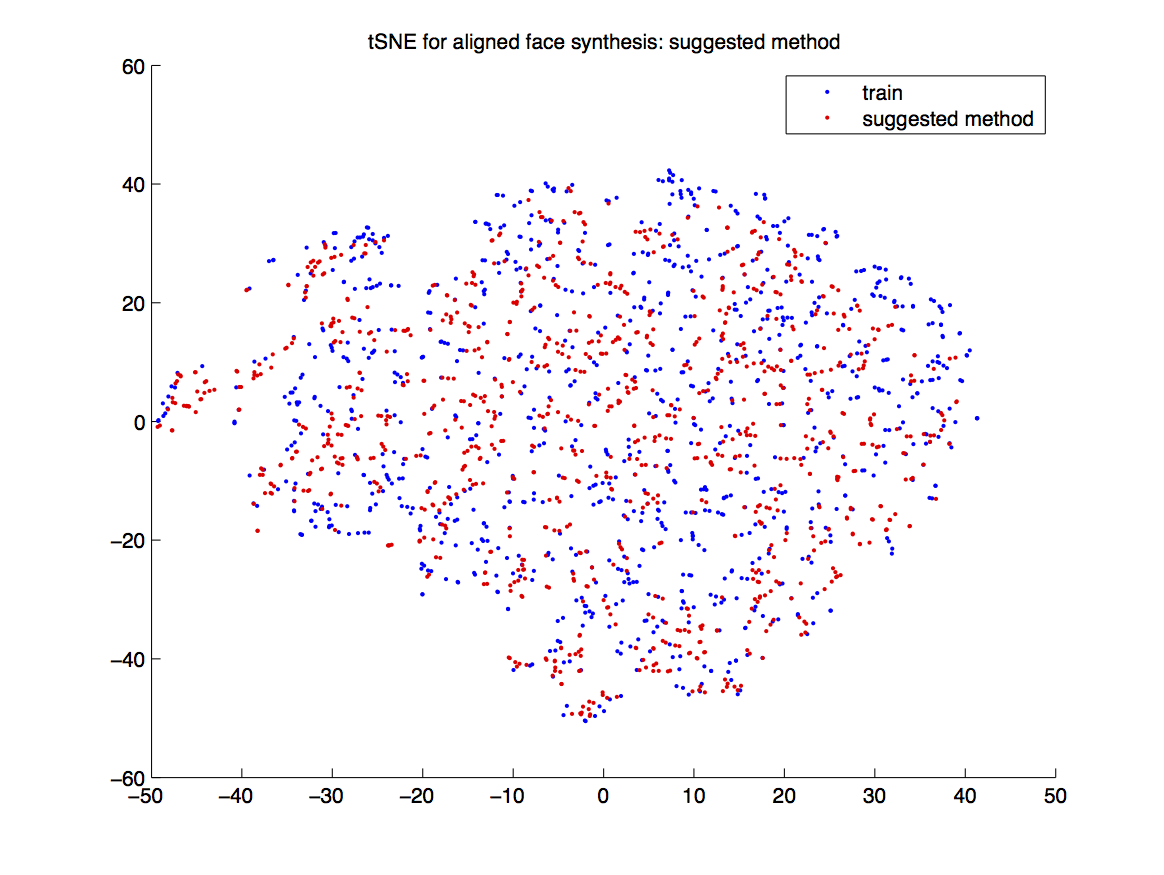}
    \caption{t-SNE visualization result of the generated aligned faces, together with part of the training set. The $x$ and $y$-axes represent the two most significant geometry components found by t-SNE.}
    \label{fig:aligned face tsne}
\end{figure}

\begin{figure}
    \centering
    \includegraphics[width=0.9\textwidth]{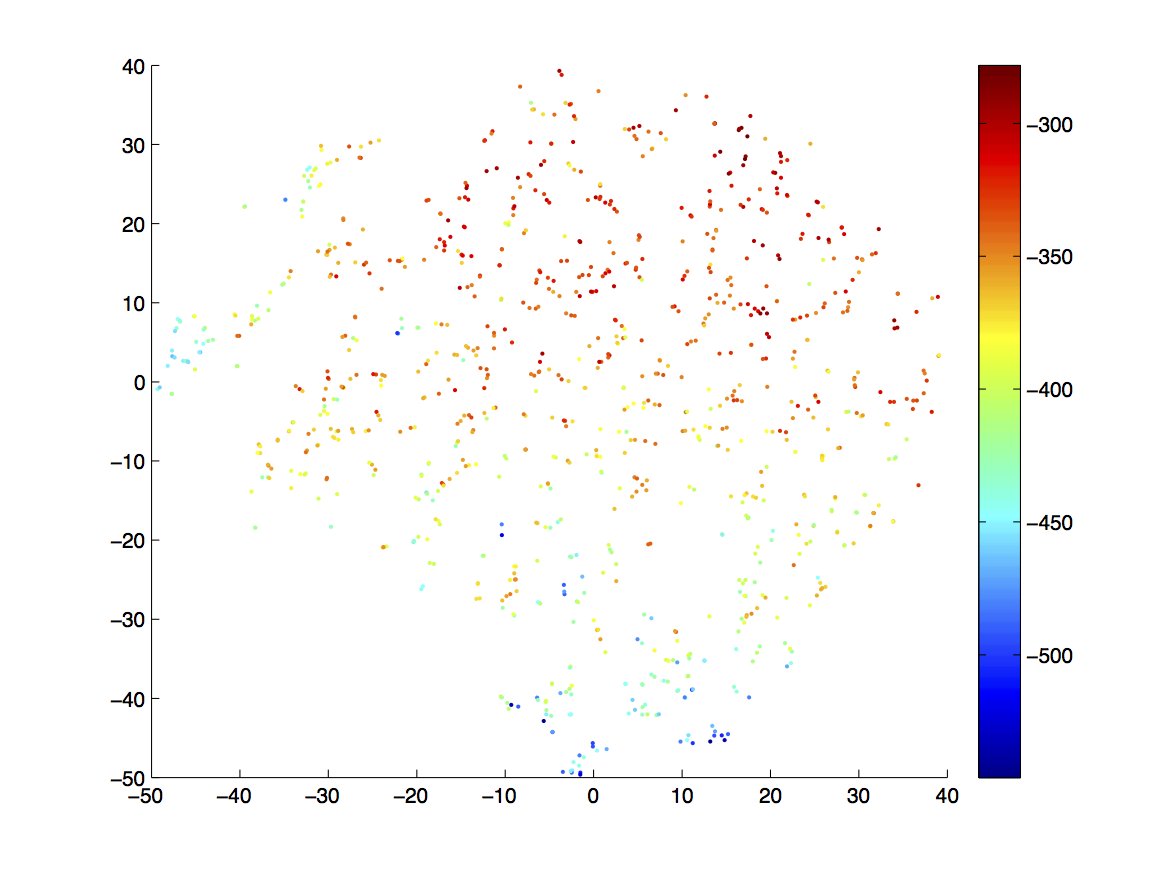}
    \caption{LL visualization of the generated aligned faces. The positions of the points are the same as the red points in Figure \ref{fig:aligned face tsne}. The colors of the points correspond to the LL value of the images.}
    \label{fig:aligned face tsne ll}
\end{figure}

\begin{figure}
    \centering
    \includegraphics[width=0.9\textwidth]{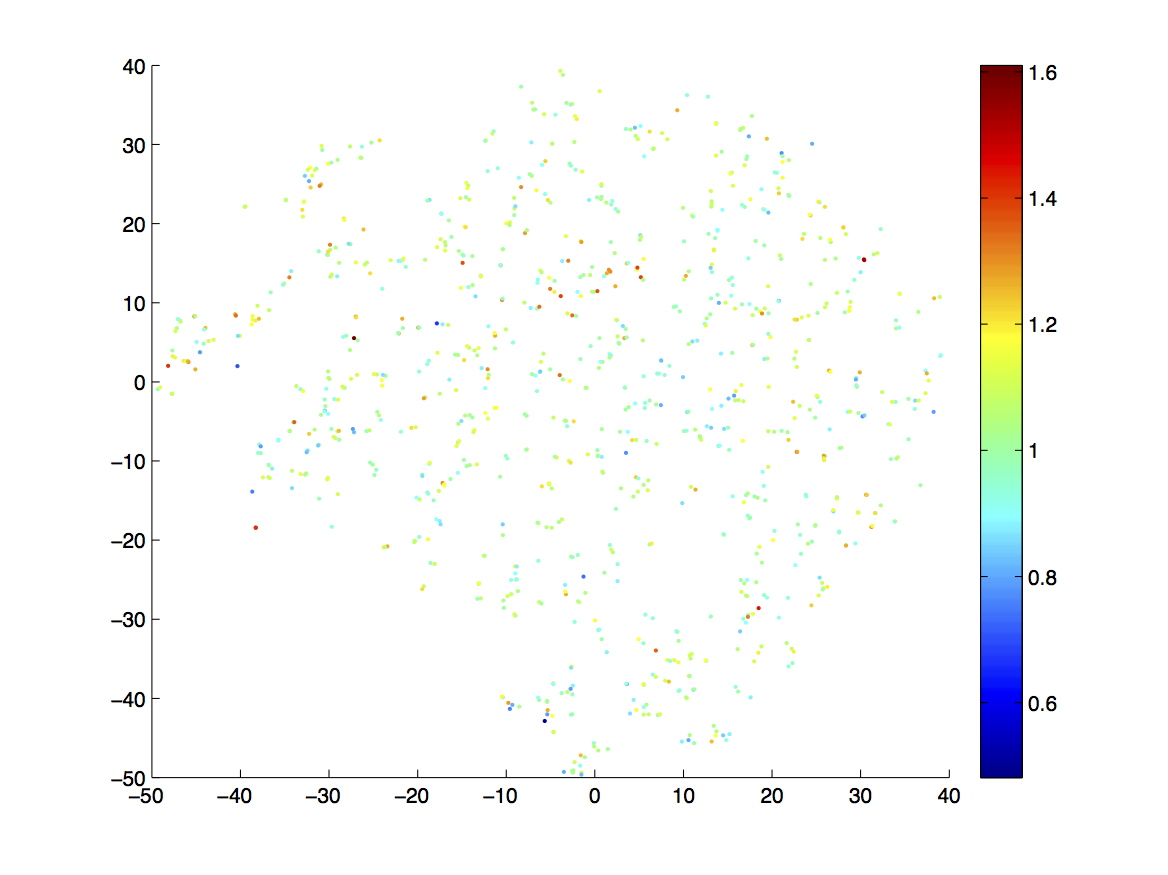}
    \caption{Originality visualization of the generated aligned faces. The positions of the points are the same as the red points in Figure \ref{fig:aligned face tsne}. The colors of the points correspond to the originality value of the images.}
    \label{fig:aligned face tsne originality}
\end{figure}

Compared to the high variability and complexity of human faces, our small face dataset is far from being enough to generate new faces with good spread. As a result, the spread assessment may not be meaningful, and we provide the results here merely for completeness. Our results has a better spread score than the test images, as can be seen in Table \ref{table:aligned faces assessment scores}. The visualization results are shown in Figures \ref{fig:aligned face tsne}, \ref{fig:aligned face tsne ll} and \ref{fig:aligned face tsne originality}. In Figure \ref{fig:aligned face tsne} the generated images spread reasonably and evenly over the training images.

% As for the spread, although our outcome has a better score, it may be due to the small number of test images. In addition, in case of the test images, most spread values are negative and far from zero, indicating that the test images are very different from the training ones (as the high originality of test images shows), whereas our images are generated based on the training set, leading to more similarities. The t-SNE result and the visualization of spread are shown in Figures  \ref{fig:aligned face tsne} and \ref{fig:aligned face spread}. In Figure \ref{fig:aligned face tsne} the generated images spread reasonably and evenly over the training images.


% As the results of non-aligned faces are not yet pleasant, we omit the assessment in this case.

\section{Discussion}
\label{sec:discussion}

The synthesis scheme we propose generates high quality MNIST images competitive to the state-of-the-art, and visually appealing face images. Furthermore, it is general since various patch priors can be easily plugged into the generation framework to create completely different kinds of images (e.g. digits and faces). Finally, our proposed prior is non-parametric and intuitive opposed to the majority of recent works, which rely on neural networks as blackboxes, thus lose traceability.

In addition, our evaluation framework makes the first step towards a complete way to assess the synthesis performance, by evaluating the LL, the originality and the spread of the generated images. The framework borrows tools from the well known t-SNE\cite{van2008visualizing}, which visualizes the MNIST digits in an impressive way.

One can wonder how other patch priors work in synthesis compared to the example-based one in Section \ref{subsec:local nn patch prior}. In fact, a natural alternative is GMM, which provides state-of-the-art restoration results when used with EPLL\cite{zoran2011learning}, and has a natural definition of patch LL. However, we believe that GMM tends to prefer smooth patches as their probability is emphasized in the Gaussian distribution, and this is a shortcoming if we are to generate sharp images. On the other hand, as we see in section \ref{subsec:assess mnist}, our current example-based model does not generate new patches, which limits the novelty the generated images can get. Therefore, it will be interesting to find a patch prior able to generate new yet sharp and likely patches.

A promising future direction is to extend our algorithm to work with more complex images, e.g. natural scenes. However, we believe that simple adaptions are not enough, as natural images have rich high-level structures, for example objects and their relations, and these images have weaker locality than digits and faces. Therefore, more delicate work may be needed towards using the proposed scheme for getting appealing natural image synthesis. One possibility is to model the patches with better use of their context, or in a feature space (e.g. as done in \cite{hertzmann2001image}).

Another future work direction consists of further improving the proposed assessment framework. Recall that two of the three measures (namely the spread and the originality) are based on the $L_2$-distance. This distance is a reasonable choice for intrinsically low dimensional images with enough samples, e.g. digits. However, when it comes to faces or even natural images, $L_2$-distance includes less information and the desired properties of natural images (e.g. invariance to shift, rotation and luminance) are largely missing in this metric. Therefore, an appropriate metric is crucial for us to assess generated complex images in a meaningful way. A recent work on synthesis with neural networks\cite{larsen2015autoencoding} suggests trained metrics, showing one possibility of achieving this goal.

\section*{Acknowledgements}
\addcontentsline{toc}{section}{Acknowledgements}

The research leading to these results has received funding from the European Research Council under European Unions Seventh Framework Programme, ERC Grant agreement no. 320649.

\appendix
\section{Choice of $\sigma$ in Log-likelihood}
\label{appendix:sigma}
In Section \ref{subsec:ll} we note that the rank estimation of the local covariance matrix is crucial for the numerical computation of LL measure. Specifically, as a way to bypass the rank estimation we choose the special value $\sigma = 1 / \sqrt{2\pi}$, so that the rank has no effect to the LL value. In this section, we show the average LL value of one specific kind of MNIST digits -- the digit 5 -- as a function of $\sigma$ using a simple rank estimation technique. This provides an evidence that $\sigma = 1 / \sqrt{2\pi}$ is a reasonable choice in case of our digit and face assessment.

The basic idea of taking the rank into account when evaluating the LL of a test patch $x$, is to construct a covariance matrix $\Sigma_{\underline{y}_j}$ around each training patch $\underline{y}_j \in D_{HR}^{i, 0}$ (see Equation (\ref{eq:patch ll})), and then replace the dimension of the patch $n^2$ by the numerical rank of $\Sigma_{\underline{y}_j}$ for each $\underline{y}_j$. Formally, Equation (\ref{eq:patch ll}) is transformed to
\begin{equation} \label{eq: patch pb with rank}
P_i (\underline{x}) = \frac{1}{\abs{D_{HR}^{i, 0}}} \sum_{\underline{y}_j \in D_{HR}^{i, 0}} \frac{1}{(2\pi \sigma^2)^{d_{\underline{y}_j}/2}} \exp\left\{\frac{-\norm{\underline{x} -\underline{y}_j}^2}{2\sigma^2}\right\}.
\end{equation}
In the above,
$$
d_{\underline{y}_j} = \rank(\Sigma_{\underline{y}_j}, \epsilon), \quad \forall \underline{y}_j \in D_{HR}^{i, 0},
$$
where $\epsilon$ is the tolerance of the numerical rank, i.e. the rank is defined by the number of eigenvalues greater than $\epsilon$. Furthermore,
\begin{align*}
    \Sigma_{\underline{y}_j} &= \sum_{\underline{z} \in D_{HR}^{i, 0}} p_{\underline{z}|\underline{y}_j} (\underline{z} - \underline{y}_j)^T (\underline{z} - \underline{y}_j), \\
    p_{\underline{z}|\underline{y}_j} &= \frac{\exp(-\norm{\underline{y}_j - \underline{z}}_2^2 / 2\sigma_i^2)}{\sum_{\underline{w} \in D_{HR}^{i, 0}, \underline{w} \neq \underline{y}_j} \exp(-\norm{\underline{y}_j - \underline{w}}_2^2 / 2\sigma_i^2)}
\end{align*}
We define $p_{\underline{z}|\underline{y}_j}$ in a fashion similar to that of the t-SNE (see Section \ref{subsec:spread}) for robustness. However, a simpler version can be suggested by using the $k$ nearest neighbors search and setting $p_{\underline{z}|\underline{y}_j}$ to be equal to $1/k$ for a neighboring patch and zero otherwise.

\begin{figure}
    \centering
    \includegraphics[width=0.9\textwidth]{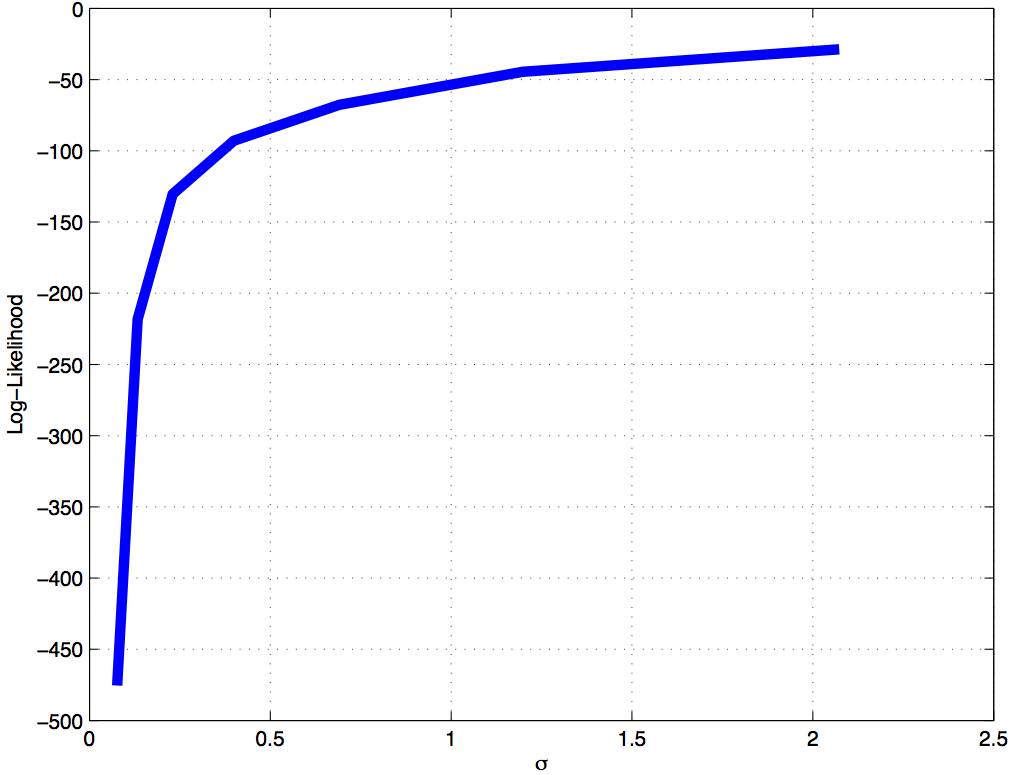}
    \caption{The average LL of digits ``5" from the MNIST test set as a function of $\sigma$. The tolerance for numerical rank is $0.1$.}
    \label{fig:sigma ll}
\end{figure}

\begin{figure}
    \centering
    \includegraphics[width=0.9\textwidth]{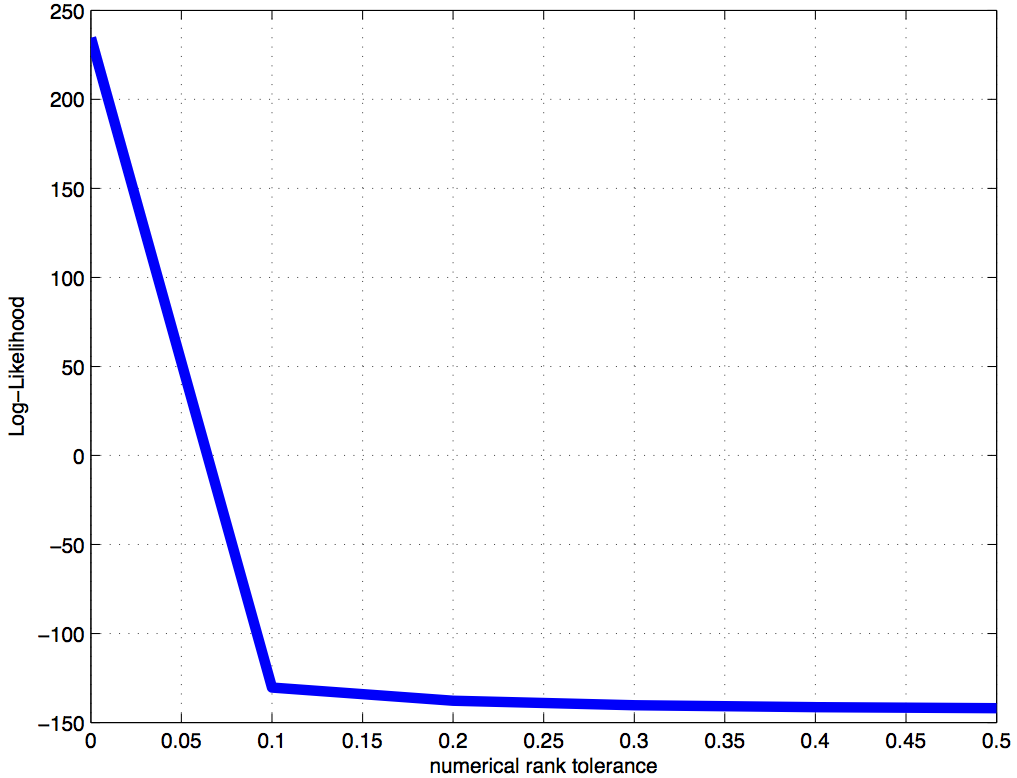}
    \caption{The average LL of digits ``5" from the MNIST test set as a function of the tolerance of numerical rank. $\sigma = 1 / \sqrt{6\pi}$.}
    \label{fig:tolerance ll}
\end{figure}

Using $P_i (\underline{x})$ defined in Equation (\ref{eq: patch pb with rank}), we evaluate the average LL values of the digits ``5" from the MNIST test set. Figure \ref{fig:sigma ll} shows how the LL varies as a function of $\sigma$ (with $\epsilon = 0.1$), and Figure \ref{fig:tolerance ll} illustrates the effect of $\epsilon$ on the LL measure. As can be seen, different values of $\sigma$ and $\epsilon$ lead to different LL scores. This sheds light on our chosen value of $\sigma = 1 / \sqrt{2\pi}$ which is independent of the rank estimation and therefore maintains the neutrality of the LL measure.

For completeness, notice that it is possible to generalize $P_i(\underline{x})$ by replacing its covariance matrix $\sigma I$ by $\Sigma_{\underline{y}_j}$ for each $\underline{y}_j \in D_{HR}^{i, 0}$, and $P_i(\underline{x})$ becomes
$$
P_i (\underline{x}) = \frac{1}{\abs{D_{HR}^{i, 0}}} \sum_{\underline{y}_j \in D_{HR}^{i, 0}} (\abs{2\pi \Sigma_{\underline{y}_j}}_+)^{-1/2} \exp\left\{-\frac{1}{2}(\underline{x}-\underline{y}_j)^T \Sigma_{\underline{y}_j}^{+}(\underline{x}-\underline{y}_j)\right\},
$$
where $\abs{2\pi \Sigma_{\underline{y}_j}}_+$ is the pseudo-determinant of $2\pi \Sigma_{\underline{y}_j}$. However, the above $P_i(\underline{x})$ is computationally prohibitive and also suffers from numerical issues especially when evaluating the determinant and the pseudo-inverse of $\Sigma_{\underline{y}_j}$. Therefore we choose to avoid this option as well.

% \section{\cmt{Things that may be added}}

% \cmt{Compare spatially variant model with spatially invariant model.

% Compare local models with and without context.

% Discuss effects of the parameters and compare results: window size, number of iterations, overlaps, $h$, $\rho$ etc.

% Update literature survey.

% Test robustness of RBM LL and our LL on bad images. (we may have the same issues as RBM)

% Show more generation results.

% Aligned face: better visual quality, find out where there are grey areas in hair.}

\bibliographystyle{IEEEtran}
\bibliography{references}

\begin{thebibliography}{10}
\providecommand{\url}[1]{#1}
\csname url@samestyle\endcsname
\providecommand{\newblock}{\relax}
\providecommand{\bibinfo}[2]{#2}
\providecommand{\BIBentrySTDinterwordspacing}{\spaceskip=0pt\relax}
\providecommand{\BIBentryALTinterwordstretchfactor}{4}
\providecommand{\BIBentryALTinterwordspacing}{\spaceskip=\fontdimen2\font plus
\BIBentryALTinterwordstretchfactor\fontdimen3\font minus
  \fontdimen4\font\relax}
\providecommand{\BIBforeignlanguage}[2]{{%
\expandafter\ifx\csname l@#1\endcsname\relax
\typeout{** WARNING: IEEEtran.bst: No hyphenation pattern has been}%
\typeout{** loaded for the language `#1'. Using the pattern for}%
\typeout{** the default language instead.}%
\else
\language=\csname l@#1\endcsname
\fi
#2}}
\providecommand{\BIBdecl}{\relax}
\BIBdecl

\bibitem{elad2006image}
M.~Elad and M.~Aharon, ``{Image denoising via sparse and redundant
  representations over learned dictionaries},'' \emph{Image Processing, IEEE
  Transactions on}, vol.~15, no.~12, pp. 3736--3745, 2006.

\bibitem{dabov2007image}
K.~Dabov, A.~Foi, V.~Katkovnik, and K.~Egiazarian, ``{Image denoising by sparse
  3-D transform-domain collaborative filtering},'' \emph{Image Processing, IEEE
  Transactions on}, vol.~16, no.~8, pp. 2080--2095, 2007.

\bibitem{mairal2008sparse}
J.~Mairal, M.~Elad, and G.~Sapiro, ``{Sparse representation for color image
  restoration},'' \emph{Image Processing, IEEE Transactions on}, vol.~17,
  no.~1, pp. 53--69, 2008.

\bibitem{criminisi2004region}
A.~Criminisi, P.~P{\'e}rez, and K.~Toyama, ``{Region filling and object removal
  by exemplar-based image inpainting},'' \emph{Image Processing, IEEE
  Transactions on}, vol.~13, no.~9, pp. 1200--1212, 2004.

\bibitem{yang2010image}
J.~Yang, J.~Wright, T.~S. Huang, and Y.~Ma, ``{Image super-resolution via
  sparse representation},'' \emph{Image Processing, IEEE Transactions on},
  vol.~19, no.~11, pp. 2861--2873, 2010.

\bibitem{zoran2011learning}
D.~Zoran and Y.~Weiss, ``{From learning models of natural image patches to
  whole image restoration},'' in \emph{Computer Vision (ICCV), 2011 IEEE
  International Conference on}.\hskip 1em plus 0.5em minus 0.4em\relax IEEE,
  2011, pp. 479--486.

\bibitem{hertzmann2001image}
A.~Hertzmann, C.~E. Jacobs, N.~Oliver, B.~Curless, and D.~H. Salesin, ``{Image
  analogies},'' in \emph{Proceedings of the 28th annual conference on Computer
  graphics and interactive techniques}.\hskip 1em plus 0.5em minus 0.4em\relax
  ACM, 2001, pp. 327--340.

\bibitem{cheng2008consistent}
L.~Cheng, S.~N. Vishwanathan, and X.~Zhang, ``{Consistent image analogies using
  semi-supervised learning},'' in \emph{Computer Vision and Pattern
  Recognition, 2008. CVPR 2008. IEEE Conference on}.\hskip 1em plus 0.5em minus
  0.4em\relax IEEE, 2008, pp. 1--8.

\bibitem{benard2013stylizing}
P.~B{\'e}nard, F.~Cole, M.~Kass, I.~Mordatch, J.~Hegarty, M.~S. Senn,
  K.~Fleischer, D.~Pesare, and K.~Breeden, ``{Stylizing animation by
  example},'' \emph{ACM Transactions on Graphics (TOG)}, vol.~32, no.~4, p.
  119, 2013.

\bibitem{barnes2015patchtable}
C.~Barnes, F.-L. Zhang, L.~Lou, X.~Wu, and S.-M. Hu, ``{PatchTable: efficient
  patch queries for large datasets and applications},'' \emph{ACM Transactions
  on Graphics (TOG)}, vol.~34, no.~4, p.~97, 2015.

\bibitem{kyprianidis2013state}
J.~E. Kyprianidis, J.~Collomosse, T.~Wang, and T.~Isenberg, ``{State of the \lq
  Art\rq: A taxonomy of artistic stylization techniques for images and
  video},'' \emph{IEEE Transactions on Visualization and Computer Graphics},
  vol.~19, no.~5, pp. 866--885, 2013.

\bibitem{gatys2016image}
L.~A. Gatys, A.~S. Ecker, and M.~Bethge, ``{Image style transfer using
  convolutional neural networks},'' in \emph{Proceedings of the IEEE Conference
  on Computer Vision and Pattern Recognition}, 2016, pp. 2414--2423.

\bibitem{johnson2016perceptual}
J.~Johnson, A.~Alahi, and L.~Fei-Fei, ``{Perceptual losses for real-time style
  transfer and super-resolution},'' \emph{arXiv preprint arXiv:1603.08155},
  2016.

\bibitem{bruna2015super}
J.~Bruna, P.~Sprechmann, and Y.~LeCun, ``{Super-resolution with deep
  convolutional sufficient statistics},'' \emph{arXiv preprint
  arXiv:1511.05666}, 2015.

\bibitem{dong2016image}
C.~Dong, C.~C. Loy, K.~He, and X.~Tang, ``{Image super-resolution using deep
  convolutional networks},'' \emph{IEEE transactions on pattern analysis and
  machine intelligence}, vol.~38, no.~2, pp. 295--307, 2016.

\bibitem{radford2015unsupervised}
A.~Radford, L.~Metz, and S.~Chintala, ``{Unsupervised representation learning
  with deep convolutional generative adversarial networks},'' \emph{arXiv
  preprint arXiv:1511.06434}, 2015.

\bibitem{van2016pixel}
A.~van~den Oord, N.~Kalchbrenner, and K.~Kavukcuoglu, ``{Pixel recurrent neural
  networks},'' \emph{arXiv preprint arXiv:1601.06759}, 2016.

\bibitem{gregor2015draw}
K.~Gregor, I.~Danihelka, A.~Graves, D.~Rezende, and D.~Wierstra, ``{DRAW: A
  Recurrent neural network for image generation},'' in \emph{Proceedings of The
  32nd International Conference on Machine Learning}, 2015, pp. 1462--1471.

\bibitem{kingma2016improving}
D.~P. Kingma, T.~Salimans, and M.~Welling, ``{Improving variational inference
  with inverse autoregressive flow},'' \emph{arXiv preprint arXiv:1606.04934},
  2016.

\bibitem{salimans2016improved}
T.~Salimans, I.~Goodfellow, W.~Zaremba, V.~Cheung, A.~Radford, and X.~Chen,
  ``{Improved Techniques for Training GANs},'' \emph{arXiv preprint
  arXiv:1606.03498}, 2016.

\bibitem{google2015inceptionism}
G.~Research, ``{Inceptionism: going deeper into neural networks},'' 2015,
  accessed: 2015-08-06.

\bibitem{portilla2000parametric}
J.~Portilla and E.~P. Simoncelli, ``{A parametric texture model based on joint
  statistics of complex wavelet coefficients},'' \emph{International Journal of
  Computer Vision}, vol.~40, no.~1, pp. 49--70, 2000.

\bibitem{peyre2010texture}
G.~Peyr{\'e}, ``{Texture synthesis with grouplets},'' \emph{Pattern Analysis
  and Machine Intelligence, IEEE Transactions on}, vol.~32, no.~4, pp.
  733--746, 2010.

\bibitem{peyre2009sparse}
------, ``{Sparse modeling of textures},'' \emph{Journal of Mathematical
  Imaging and Vision}, vol.~34, no.~1, pp. 17--31, 2009.

\bibitem{tartavel2014variational}
G.~Tartavel, Y.~Gousseau, and G.~Peyr{\'e}, ``{Variational texture synthesis
  with sparsity and spectrum constraints},'' \emph{Journal of Mathematical
  Imaging and Vision}, vol.~52, no.~1, pp. 124--144, 2014.

\bibitem{paget1995texture}
R.~Paget and D.~Longstaff, ``{Texture synthesis via a non-parametric Markov
  random field},'' \emph{Proceedings of DICTA-95, Digital Image Computing:
  Techniques and Applications}, vol.~1, pp. 547--552, 1995.

\bibitem{efros2001image}
A.~A. Efros and W.~T. Freeman, ``{Image quilting for texture synthesis and
  transfer},'' in \emph{Proceedings of the 28th annual conference on Computer
  graphics and interactive techniques}.\hskip 1em plus 0.5em minus 0.4em\relax
  ACM, 2001, pp. 341--346.

\bibitem{kwatra2003graphcut}
V.~Kwatra, A.~Sch{\"o}dl, I.~Essa, G.~Turk, and A.~Bobick, ``{Graphcut
  textures: Image and video synthesis using graph cuts},'' in \emph{ACM
  Transactions on Graphics (ToG)}, vol.~22, no.~3.\hskip 1em plus 0.5em minus
  0.4em\relax ACM, 2003, pp. 277--286.

\bibitem{kwatra2005texture}
V.~Kwatra, I.~Essa, A.~Bobick, and N.~Kwatra, ``{Texture optimization for
  example-based synthesis},'' \emph{ACM Transactions on Graphics (ToG)},
  vol.~24, no.~3, pp. 795--802, 2005.

\bibitem{gatys2015texture}
L.~Gatys, A.~S. Ecker, and M.~Bethge, ``{Texture synthesis using convolutional
  neural networks},'' in \emph{Advances in Neural Information Processing
  Systems}, 2015, pp. 262--270.

\bibitem{ashikhmin2001synthesizing}
M.~Ashikhmin, ``{Synthesizing natural textures},'' in \emph{Proceedings of the
  2001 symposium on Interactive 3D graphics}.\hskip 1em plus 0.5em minus
  0.4em\relax ACM, 2001, pp. 217--226.

\bibitem{lefebvre2005parallel}
S.~Lefebvre and H.~Hoppe, ``{Parallel controllable texture synthesis},'' in
  \emph{ACM Transactions on Graphics (ToG)}, vol.~24, no.~3.\hskip 1em plus
  0.5em minus 0.4em\relax ACM, 2005, pp. 777--786.

\bibitem{lefebvre2006appearance}
------, ``{Appearance-space texture synthesis},'' \emph{ACM Transactions on
  Graphics (TOG)}, vol.~25, no.~3, pp. 541--548, 2006.

\bibitem{efros1999texture}
A.~Efros, T.~K. Leung \emph{et~al.}, ``{Texture synthesis by non-parametric
  sampling},'' in \emph{Computer Vision, 1999. The Proceedings of the Seventh
  IEEE International Conference on}, vol.~2.\hskip 1em plus 0.5em minus
  0.4em\relax IEEE, 1999, pp. 1033--1038.

\bibitem{salakhutdinov2008quantitative}
R.~Salakhutdinov and I.~Murray, ``{On the quantitative analysis of deep belief
  networks},'' in \emph{Proceedings of the 25th international conference on
  Machine learning}.\hskip 1em plus 0.5em minus 0.4em\relax ACM, 2008, pp.
  872--879.

\bibitem{uria2014deep}
B.~Uria, I.~Murray, and H.~Larochelle, ``{A deep and tractable density
  estimator},'' in \emph{Proceedings of The 31st International Conference on
  Machine Learning}, 2014, pp. 467--475.

\bibitem{raiko2014iterative}
T.~Raiko, Y.~Li, K.~Cho, and Y.~Bengio, ``{Iterative neural autoregressive
  distribution estimator (NADE-k)},'' in \emph{Advances in Neural Information
  Processing Systems}, 2014, pp. 325--333.

\bibitem{salakhutdinov2009deep}
R.~Salakhutdinov and G.~E. Hinton, ``{Deep boltzmann machines},'' in
  \emph{International Conference on Artificial Intelligence and Statistics},
  2009, pp. 448--455.

\bibitem{murray2009evaluating}
I.~Murray and R.~R. Salakhutdinov, ``{Evaluating probabilities under
  high-dimensional latent variable models},'' in \emph{Advances in neural
  information processing systems}, 2009, pp. 1137--1144.

\bibitem{rezende2014stochastic}
D.~J. Rezende, S.~Mohamed, and D.~Wierstra, ``{Stochastic backpropagation and
  approximate inference in deep generative models},'' in \emph{Proceedings of
  The 31st International Conference on Machine Learning}, 2014, pp. 1278--1286.

\bibitem{salimans2015markov}
T.~Salimans, D.~P. Kingma, M.~Welling \emph{et~al.}, ``{Markov chain Monte
  Carlo and variational inference: Bridging the gap},'' in \emph{International
  Conference on Machine Learning}, 2015, pp. 1218--1226.

\bibitem{gregor2014deep}
K.~Gregor, I.~Danihelka, A.~Mnih, C.~Blundell, and D.~Wierstra, ``{Deep
  AutoRegressive networks},'' in \emph{Proceedings of The 31st International
  Conference on Machine Learning}, 2014, pp. 1242--1250.

\bibitem{hinton2006fast}
G.~E. Hinton, S.~Osindero, and Y.-W. Teh, ``{A fast learning algorithm for deep
  belief nets},'' \emph{Neural computation}, vol.~18, no.~7, pp. 1527--1554,
  2006.

\bibitem{goodfellow2014generative}
I.~Goodfellow, J.~Pouget-Abadie, M.~Mirza, B.~Xu, D.~Warde-Farley, S.~Ozair,
  A.~Courville, and Y.~Bengio, ``{Generative adversarial nets},'' in
  \emph{Advances in Neural Information Processing Systems}, 2014, pp.
  2672--2680.

\bibitem{kim2016deep}
T.~Kim and Y.~Bengio, ``{Deep directed generative models with energy-based
  probability estimation},'' \emph{arXiv preprint arXiv:1606.03439}, 2016.

\bibitem{germain2015made}
M.~Germain, K.~Gregor, I.~Murray, and H.~Larochelle, ``{MADE: masked
  autoencoder for distribution estimation},'' in \emph{International Conference
  on Machine Learning}, 2015, pp. 881--889.

\bibitem{ranzato2013modeling}
M.~Ranzato, V.~Mnih, J.~M. Susskind, and G.~E. Hinton, ``{Modeling natural
  images using gated MRFs},'' \emph{Pattern Analysis and Machine Intelligence,
  IEEE Transactions on}, vol.~35, no.~9, pp. 2206--2222, 2013.

\bibitem{larsen2015autoencoding}
A.~B.~L. Larsen, S.~K. S{\o}nderby, and O.~Winther, ``{Autoencoding beyond
  pixels using a learned similarity metric},'' \emph{arXiv preprint
  arXiv:1512.09300}, 2015.

\bibitem{liu2007face}
C.~Liu, H.-Y. Shum, and W.~T. Freeman, ``{Face hallucination: Theory and
  practice},'' \emph{International Journal of Computer Vision}, vol.~75, no.~1,
  pp. 115--134, 2007.

\bibitem{denton2015deep}
E.~L. Denton, S.~Chintala, R.~Fergus \emph{et~al.}, ``{Deep generative image
  models using a￼ Laplacian pyramid of adversarial networks},'' in
  \emph{Advances in neural information processing systems}, 2015, pp.
  1486--1494.

\bibitem{mahendran2015understanding}
A.~Mahendran and A.~Vedaldi, ``{Understanding deep image representations by
  inverting them},'' in \emph{2015 IEEE conference on computer vision and
  pattern recognition (CVPR)}.\hskip 1em plus 0.5em minus 0.4em\relax IEEE,
  2015, pp. 5188--5196.

\bibitem{simonyan2013deep}
K.~Simonyan, A.~Vedaldi, and A.~Zisserman, ``{Deep inside convolutional
  networks: Visualising image classification models and saliency maps},''
  \emph{arXiv preprint arXiv:1312.6034}, 2013.

\bibitem{dosovitskiy2015inverting}
A.~Dosovitskiy and T.~Brox, ``{Inverting convolutional networks with
  convolutional networks},'' \emph{CoRR abs/1506.02753}, 2015.

\bibitem{nguyen2015deep}
A.~Nguyen, J.~Yosinski, and J.~Clune, ``{Deep neural networks are easily
  fooled: High confidence predictions for unrecognizable images},'' in
  \emph{2015 IEEE Conference on Computer Vision and Pattern Recognition
  (CVPR)}.\hskip 1em plus 0.5em minus 0.4em\relax IEEE, 2015, pp. 427--436.

\bibitem{freeman2002example}
W.~T. Freeman, T.~R. Jones, and E.~C. Pasztor, ``{Example-based
  super-resolution},'' \emph{IEEE Computer graphics and Applications}, vol.~22,
  no.~2, pp. 56--65, 2002.

\bibitem{freeman2000learning}
W.~T. Freeman, E.~C. Pasztor, and O.~T. Carmichael, ``{Learning low-level
  vision},'' \emph{International journal of computer vision}, vol.~40, no.~1,
  pp. 25--47, 2000.

\bibitem{yu2012solving}
G.~Yu, G.~Sapiro, and S.~Mallat, ``{Solving inverse problems with piecewise
  linear estimators: From Gaussian mixture models to structured sparsity},''
  \emph{Image Processing, IEEE Transactions on}, vol.~21, no.~5, pp.
  2481--2499, 2012.

\bibitem{mairal2009non}
J.~Mairal, F.~Bach, J.~Ponce, G.~Sapiro, and A.~Zisserman, ``{Non-local sparse
  models for image restoration},'' in \emph{2009 IEEE 12th International
  Conference on Computer Vision}.\hskip 1em plus 0.5em minus 0.4em\relax IEEE,
  2009, pp. 2272--2279.

\bibitem{glasner2009super}
D.~Glasner, S.~Bagon, and M.~Irani, ``{Super-resolution from a single image},''
  in \emph{2009 IEEE 12th International Conference on Computer Vision}.\hskip
  1em plus 0.5em minus 0.4em\relax IEEE, 2009, pp. 349--356.

\bibitem{timofte2014a+}
R.~Timofte, V.~De~Smet, and L.~Van~Gool, ``{A+: Adjusted anchored neighborhood
  regression for fast super-resolution},'' in \emph{Asian Conference on
  Computer Vision}.\hskip 1em plus 0.5em minus 0.4em\relax Springer, 2014, pp.
  111--126.

\bibitem{papyan2016multi}
V.~Papyan and M.~Elad, ``{Multi-scale patch-based image restoration},''
  \emph{IEEE Transactions on Image Processing}, vol.~25, no.~1, pp. 249--261,
  2016.

\bibitem{romano2014single}
Y.~Romano, M.~Protter, and M.~Elad, ``{Single image interpolation via adaptive
  nonlocal sparsity-based modeling},'' \emph{IEEE Transactions on Image
  Processing}, vol.~23, no.~7, pp. 3085--3098, 2014.

\bibitem{lecun1998gradient}
Y.~LeCun, L.~Bottou, Y.~Bengio, and P.~Haffner, ``{Gradient-based learning
  applied to document recognition},'' \emph{Proceedings of the IEEE}, vol.~86,
  no.~11, pp. 2278--2324, 1998.

\bibitem{boyd2011distributed}
S.~Boyd, N.~Parikh, E.~Chu, B.~Peleato, and J.~Eckstein, ``{Distributed
  optimization and statistical learning via the alternating direction method of
  multipliers},'' \emph{Foundations and Trends{\textregistered} in Machine
  Learning}, vol.~3, no.~1, pp. 1--122, 2011.

\bibitem{aharon2006k}
M.~Aharon, M.~Elad, and A.~Bruckstein, ``{K-SVD: An Algorithm for Designing
  Overcomplete Dictionaries for Sparse Representation},'' \emph{IEEE
  Transactions on signal processing}, vol.~54, no.~11, pp. 4311--4322, 2006.

\bibitem{coifman1995wavelets}
R.~Coifman and D.~Donoho, ``{Wavelets and statistics},'' \emph{Lecture notes in
  Statistics. In: Translation-invariant de-noising. Antoniadis, A. and
  Oppenheim, G.(eds.)}, pp. 125--150, 1995.

\bibitem{ram2013image}
I.~Ram, M.~Elad, and I.~Cohen, ``{Image processing using smooth ordering of its
  patches},'' \emph{IEEE transactions on image processing}, vol.~22, no.~7, pp.
  2764--2774, 2013.

\bibitem{forsyth2012computer}
\BIBentryALTinterwordspacing
D.~Forsyth and J.~Ponce, \emph{{Computer vision: A modern approach}}, ser.
  Always learning.\hskip 1em plus 0.5em minus 0.4em\relax Pearson, 2012.
  [Online]. Available: \url{https://books.google.com/books?id=gM63QQAACAAJ}
\BIBentrySTDinterwordspacing

\bibitem{barnes2009patchmatch}
C.~Barnes, E.~Shechtman, A.~Finkelstein, and D.~Goldman, ``{PatchMatch: a
  randomized correspondence algorithm for structural image editing},''
  \emph{ACM Transactions on Graphics-TOG}, vol.~28, no.~3, p.~24, 2009.

\bibitem{ben2007gray}
G.~Ben-Artzi, H.~Hel-Or, and Y.~Hel-Or, ``{The gray-code filter kernels},''
  \emph{IEEE transactions on pattern analysis and machine intelligence},
  vol.~29, no.~3, pp. 382--393, 2007.

\bibitem{muja2009fast}
M.~Muja and D.~G. Lowe, ``{Fast approximate nearest neighbors with automatic
  algorithm configuration},'' \emph{VISAPP (1)}, vol.~2, no. 331-340, p.~2,
  2009.

\bibitem{barnes2010generalized}
C.~Barnes, E.~Shechtman, D.~B. Goldman, and A.~Finkelstein, ``{The generalized
  PatchMatch correspondence algorithm},'' in \emph{European Conference on
  Computer Vision}.\hskip 1em plus 0.5em minus 0.4em\relax Springer, 2010, pp.
  29--43.

\bibitem{xiao2011fast}
C.~Xiao, M.~Liu, N.~Yongwei, and Z.~Dong, ``{Fast exact nearest patch matching
  for patch-based image editing and processing},'' \emph{IEEE Transactions on
  Visualization and Computer Graphics}, vol.~17, no.~8, pp. 1122--1134, 2011.

\bibitem{he2012computing}
K.~He and J.~Sun, ``{Computing nearest-neighbor fields via propagation-assisted
  KD-trees},'' in \emph{Computer Vision and Pattern Recognition (CVPR), 2012
  IEEE Conference on}.\hskip 1em plus 0.5em minus 0.4em\relax IEEE, 2012, pp.
  111--118.

\bibitem{olonetsky2012treecann}
I.~Olonetsky and S.~Avidan, ``{TreeCANN -- k-d tree coherence approximate
  nearest neighbor algorithm},'' in \emph{European Conference on Computer
  Vision}.\hskip 1em plus 0.5em minus 0.4em\relax Springer, 2012, pp. 602--615.

\bibitem{romano2016patch}
Y.~Romano and M.~Elad, ``{Con-Patch: When a Patch Meets its Context},''
  \emph{Submitted to IEEE Transactions on Image Processing}, 2016.

\bibitem{jang2016tensorflowdraw}
E.~Jang, ``{TensorFlow Implementation of \lq DRAW: A Recurrent neural network
  For image generation \rq},'' \url{https://github.com/ericjang/draw}, 2016,
  accessed: 2016-08-02.

\bibitem{bryt2008compression}
O.~Bryt and M.~Elad, ``{Compression of facial images using the K-SVD
  algorithm},'' \emph{Journal of Visual Communication and Image
  Representation}, vol.~19, no.~4, pp. 270--282, 2008.

\bibitem{ram2014facial}
I.~Ram, I.~Cohen, and M.~Elad, ``{Facial image compression using
  patch-ordering-based adaptive wavelet transform},'' \emph{IEEE Signal
  Processing Letters}, vol.~21, no.~10, pp. 1270--1274, 2014.

\bibitem{elad2007low}
M.~Elad, R.~Goldenberg, and R.~Kimmel, ``{Low bit-rate compression of facial
  images},'' \emph{IEEE Transactions on Image Processing}, vol.~16, no.~9, pp.
  2379--2383, 2007.

\bibitem{parzen1962estimation}
E.~Parzen, ``{On estimation of a probability density function and mode},''
  \emph{The annals of mathematical statistics}, vol.~33, no.~3, pp. 1065--1076,
  1962.

\bibitem{van2008visualizing}
L.~Van~der Maaten and G.~Hinton, ``{Visualizing data using t-SNE},''
  \emph{Journal of Machine Learning Research}, vol.~9, no. 2579-2605, p.~85,
  2008.

\bibitem{tenenbaum2000global}
J.~B. Tenenbaum, V.~De~Silva, and J.~C. Langford, ``{A global geometric
  framework for nonlinear dimensionality reduction},'' \emph{science}, vol.
  290, no. 5500, pp. 2319--2323, 2000.

\bibitem{roweis2000nonlinear}
S.~T. Roweis and L.~K. Saul, ``{Nonlinear dimensionality reduction by locally
  linear embedding},'' \emph{Science}, vol. 290, no. 5500, pp. 2323--2326,
  2000.

\end{thebibliography}

\end{document}